\documentclass[oneside, 11pt]{Latex/Classes/PhDthesisPSnPDF}

\usepackage[T1]{fontenc}
\usepackage{array}
\usepackage{pdfpages}
\usepackage{amsmath}
\usepackage{amssymb}
\usepackage{algorithm}
\usepackage{algpseudocode}
\usepackage{booktabs}
\usepackage{amsfonts}
\usepackage{amssymb}
\usepackage{adjustbox}
\usepackage{graphicx}

\textwidth
15cm
\textheight
22cm
\parindent
10pt
\oddsidemargin
0.85cm
\evensidemargin
0.37cm

\title{Bridging Synthetic and Real-World Domains: A Human-in-the-Loop Weakly-Supervised
Framework for Industrial Toxic Emission Segmentation}
\author{%
\begin{tabular}{c@{\hskip 2cm}c}
  Yida Tao                         & Yen-Chia Hsu               \\
  Universiteit van Amsterdam       & Universiteit van Amsterdam \\
  \texttt{yida.tao@student.uva.nl} & \texttt{y.c.hsu@uva.nl}
\end{tabular}
}

\begin{document}
  \thispagestyle{empty}
  \maketitle



  \renewcommand{\baselinestretch}{1.2}
  \baselineskip=18pt plus1pt

  \frontmatter

\begin{abstracts} 
    Industrial smoke segmentation is critical for air-quality monitoring and
    environmental protection but is often hampered by the high cost and scarcity
    of pixel-level annotations in real-world settings. We introduce CEDANet, a
    human-in-the-loop, class-aware domain adaptation framework that uniquely integrates
    weak, citizen-provided video-level labels with adversarial feature alignment.
    Specifically, we refine pseudo-labels generated by a source-trained
    segmentation model using citizen votes, and employ class-specific domain discriminators
    to transfer rich source-domain representations to the industrial domain.
    Comprehensive experiments on SMOKE5K and custom IJmond datasets demonstrate
    that CEDANet achieves an $F_{1}$-score of 0.414 and a smoke-class IoU of
    0.261 with citizen feedback, vastly outperforming the baseline model, which
    scored 0.083 and 0.043 respectively. This represents a five-fold increase in
    $F_{1}$-score and a six-fold increase in smoke-class IoU. Notably, CEDANet with
    citizen-constrained pseudo-labels achieves performance comparable to the same
    architecture trained on limited 100 fully annotated images with $F_{1}$-score
    of 0.418 and IoU of 0.264, demonstrating its ability to reach small-sampled
    fully supervised-level accuracy without target-domain annotations. Our research
    validates the scalability and cost-efficiency of combining citizen science with
    weakly supervised domain adaptation, offering a practical solution for
    complex, data-scarce environmental monitoring applications.
\end{abstracts}


  \setcounter{secnumdepth}{3} 
  \setcounter{tocdepth}{3} 
  \tableofcontents 


  \listoffigures 

  \listoftables 






\nomenclature{$\mathcal{M}$}{Represents a segmentation model (e.g., $\mathcal{M}_{DA}$, $\mathcal{M}_{pretrain}$).}
\nomenclature{$\mathcal{D}$}{Represents a dataset (e.g., $\mathcal{D}_{\mathrm{smoke5k}}$, $\mathcal{D}_{\mathrm{ijmond}}$).}
\nomenclature{$\mathcal{E}$}{Represents a specific experimental setup or configuration.}
\nomenclature{$\mathcal{V}$}{Represents a set of videos.} \nomenclature{$\mathcal{F}$}{Represents a set of frames extracted from a video.}

\nomenclature{$v$}{A video from the dataset.} \nomenclature{$f$}{A frame extracted from a video.}
\nomenclature{$x$}{Input image (tensor $\in \mathbb{R}^{3 \times H \times W}$).}
\nomenclature{$y$}{Ground truth segmentation mask (label).} \nomenclature{$G$}{Feature generator network in the domain adaptation framework.}
\nomenclature{$D$}{Domain discriminator network (e.g., $D_{smoke}$, $D_{bg}$).} \nomenclature{$\mathcal{R}$}{Gradient Reversal Layer (GRL) operator.}
\nomenclature{$z$}{Latent variable in the VAE architecture (dimension 8).} \nomenclature{$\mu, \text{logvar}$}{Mean and log-variance of the latent variable distribution.}
\nomenclature{$F$}{Feature map produced by the generator ($\in \mathbb{R}^{C \times H \times W}$).}
\nomenclature{$M$}{Class-wise mask filter applied to a feature map.} \nomenclature{$A$}{Class attention map used for attention-guided pooling ($\in \mathbb{R}^{1 \times H \times W}$).}
\nomenclature{$\epsilon$}{Random noise sampled from standard normal distribution $\mathcal{N}(0,I)$.}
\nomenclature{$H, W, C$}{Height, width, and channel dimensions of tensors.}

\nomenclature{$\mathcal{L}$}{Loss function (e.g., $\mathcal{L}_{gen}$, $\mathcal{L}_{cont}$, $\mathcal{L}_{DA}$).}
\nomenclature{$L_{D}$}{Domain classification loss for discriminator.}
\nomenclature{$\lambda$}{Hyperparameter for weighting loss terms (e.g., $\lambda_{grl}$, $\lambda_{cont}$, $\lambda_{DA}$).}
\nomenclature{$\tau$ (contrastive)}{Temperature parameter in the contrastive loss function.}
\nomenclature{$\tau$ (threshold)}{Probability threshold for binarizing smoke probability maps in pseudo-labeling.}
\nomenclature{$N_{S}, N_{T}$}{Number of samples in source and target datasets, respectively.}

\nomenclature{$P(f)$}{Pixel-wise smoke probability map for frame $f$.} \nomenclature{$C(f)$}{Confidence score for frame $f$ based on mean activation and foreground ratio.}
\nomenclature{$\hat{y}$}{Binarized segmentation mask obtained from probability map.}
\nomenclature{$\hat{Y}$}{Set of pseudo-labels generated for the target domain.} \nomenclature{$\mathrm{sim}(u, v)$}{Similarity score between two frames $u$ and $v$.}
\nomenclature{$\alpha, \beta, \gamma$}{Weighting factors in heuristic-based frame selection strategy.}
\nomenclature{$k$}{Number of top candidate frames selected in multi-stage frame selection.}
  \printnomenclature[1.5cm] \label{nom}


  \mainmatter

  \renewcommand{\chaptername}{} 




\chapter{Introduction}

\ifpdf \graphicspath{{1_introduction/figures/PNG/}{1_introduction/figures/PDF/}{1_introduction/figures/}}
\else \graphicspath{{1_introduction/figures/EPS/}{1_introduction/figures/}} \fi



\section{Context}
\label{sec:Context} Hazardous smoke poses serious threats to human health and
ecosystems~\cite{Pellegrino2025}, making its detection crucial for public safety
and environmental protection. In particular, \textbf{industrial smoke detection}
has emerged as a distinct subfield within computer vision research~\cite{Hsu-2016-5586,10337640}.
Early detection systems based on optical and ionization sensors laid important foundations~\cite{Pellegrino2025,Wang2024},
but the increasing complexity of real-world environments demands more advanced
techniques. Recent work has utilized UAV-mounted sensors and AI models to improve
detection accuracy, latency, and adaptability across diverse scenarios~\cite{BOUGUETTAYA2022108309,Easwaramoorthy2023}.

\begin{figure}[htbp]
  \centering
  \includegraphics[width=0.23\textwidth]{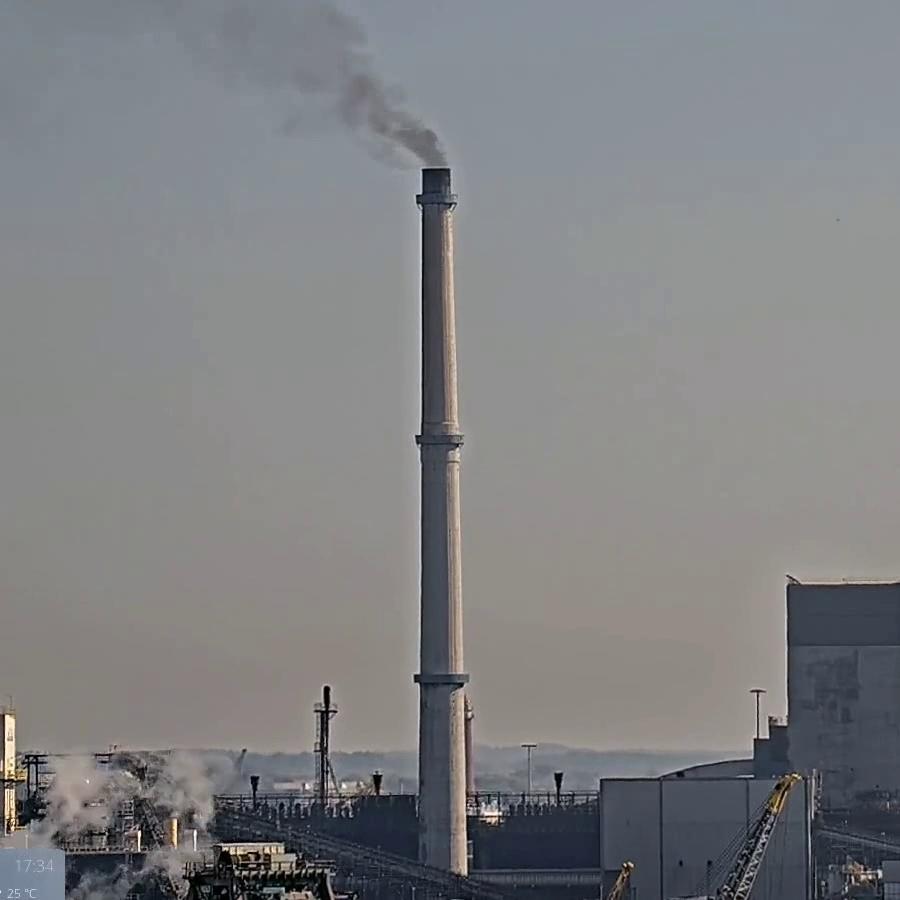}
  \includegraphics[width=0.23\textwidth]{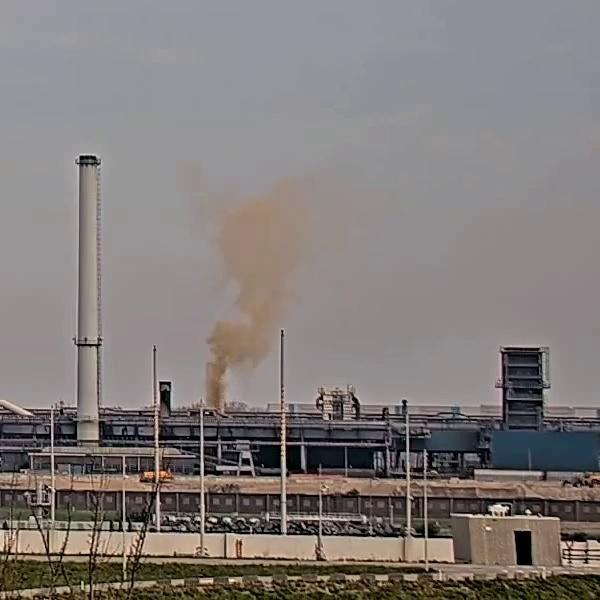}
  \hfill
  \includegraphics[width=0.23\textwidth]{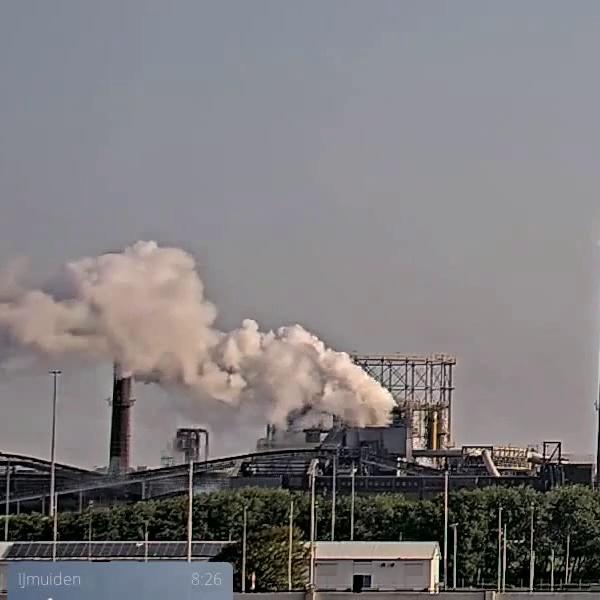}
  \includegraphics[width=0.23\textwidth]{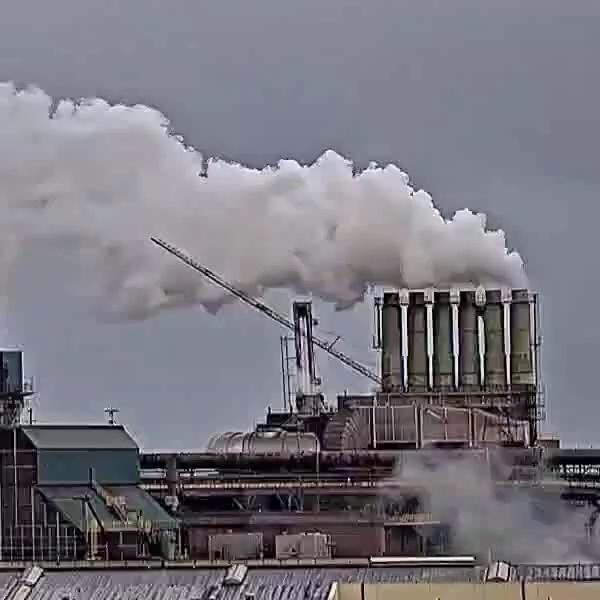}
  \caption{Emissions from factories in IJmond region, the Netherlands: the first
  two on the left show hazardous emissions, while the last two images on the right
  depict steam.}
  \label{fig:smoke-and-steam}
\end{figure}

Despite these advances, industrial smoke detection still suffers from a critical
lack of annotated data. Pixel-level labeling of industrial emissions can take up
significant human effort~\cite{peláezvegas2023surveysemisupervisedsemanticsegmentation}.
Existing smoke datasets, such as SMOKE5K, are primarily composed of synthetic
and real-world wildfire smoke images, leading to a substantial domain gap when applied
to industrial toxic cloud detection scenarios~\cite{yan2023transmissionguidedbayesiangenerativemodel}.
Industrial smoke, usually stemming from factories, power plants, and chemical facilities,
exhibits distinct visual and physical characteristics compared to natural smoke.
For instance, industrial smoke can vary widely in color, density, and texture, ranging
from dense black carbon particulate matter to different shades of gray or brown caused
by various chemicals, and generally appearing more translucent than natural
smoke~\cite{NAKHJIRI2024102504}. These intrinsic differences in composition,
density, and optical properties compared to normal steam (as seen in Figure~\ref{fig:smoke-and-steam})
make models trained on natural smoke datasets perform poorly when deployed in industrial
settings.

This severe data scarcity has motivated the creation of targeted industrial
smoke resources and specialized learning paradigms. Hsu et al.~\cite{hsu2024projectriserecognizingindustrial}
release a video-labeled dataset \textit{RISE} for factory emissions, marking an
important advance in gathering coarse labels at scale. Furthermore, weakly-supervised
learning methods exploit coarse annotations, namely, image- or video-level labels
and bounding boxes, to train segmentation models with minimal pixel-level masks,
enabling the use of large-scale datasets with limited annotations~\cite{chang2020weaklysupervisedsemanticsegmentationsubcategory,ravi2024sam2segmentimages,ke2021universalweaklysupervisedsegmentation}.

Currently, \textit{domain adaptation} has emerged as a powerful technique to mitigate
performance degradation caused by domain shifts, enabling models trained on one
domain to generalize effectively to another~\cite{paul2020domainadaptivesemanticsegmentation}.
With volunteer contributions providing non-expert annotations, \textit{crowdsourcing}
has become a viable solution to efficiently gather large-scale datasets \cite{Mosqueira-Rey2022}.
These approaches have shown promising results in various domains, including smoke
detection tasks~\cite{hsu2024projectriserecognizingindustrial, yan2023transmissionguidedbayesiangenerativemodel}.

To tackle both annotation scarcity and domain gap, our research seeks to
leverage two complementary forms of supervision:
\begin{itemize}
  \item \textbf{\emph{Full, pixel-wise masks}}: from a publicly available source
    domain dataset (e.g. synthetic and existing real-world smoke images);

  \item \textbf{\emph{Coarse, Video-level labels}}: from a custom target domain dataset
    (See \ref{sec:Dataset}), only indicating the presence/absence of smoke in a
    video, which we further transform into image-level labels.
\end{itemize}

This dual-supervision strategy lays the foundations of our proposed human-in-the-loop
domain adaptation framework.

\section{Objective and Research Questions}
\label{sec:Objective and Research Questions} The objective of this thesis is to
develop a robust and efficient industrial smoke detection system tailored to
industrial toxic cloud emission in \textbf{IJmond}, a region in the Netherlands
characterized by its industrial activities via limited annotations provided by
its local community. The system aims to study the methods of using this kind of citizen
feedback to understand its capability in this specific application context. More
concretely, our aim is to answer the following research questions:
\begin{itemize}
  \item \label{RQ1} \textbf{RQ1. How can citizen-provided weak-labels be effectively
    integrated into a class-aware domain adaptation framework?} What combination
    of model architectures best leverages weak community annotations to refine
    pixel-level masks? We will elaborate our methodologies in Section~\ref{sec:Human-in-the-Loop
    Weakly Supervised Domain Adaptation}.

  \item \label{RQ2} \textbf{RQ2. How robust is the proposed approach to varying domain
    gaps?} To what extent can it bridge visual and physical discrepancies
    between source smoke (synthetic/natural) and target industrial toxic clouds?
    We will compare our domain-adaptive model with original weakly-supervised local
    contrastive learning model in Section~\ref{sec:Quantitative Comparison}.

  \item \label{RQ3} \textbf{RQ3. How does weakly-supervised feedback from human annotators
    contribute to model performance?} Specifically, how does the classification of
    binary smoke presence at the video level provided by local communities affect
    segmentation performance, compared to no-feedback and expert annotations? We
    will address this question by conducting experiments with different pseudo-labels
    in Section~\ref{sec:Quantitative Comparison}.

  \item \label{RQ4} \textbf{RQ4. How effective is the proposed pipeline on our custom
    IJmond toxic-cloud data?} In terms of segmentation accuracy and robustness
    to noisy labels, how does the pipeline perform under varying levels of weak feedback?
    We will illustrate our experiment results in Section~\ref{sec:Qualitative
    Comparison on IJmond900 Dataset}.
\end{itemize}

By addressing these research questions, this thesis aims to deliver a human-in-the-loop
semantic segmentation solution in a weakly-supervised domain adaptation setting
that can accurately and reliably detect hazardous industrial smoke, significantly
contributing to environmental protection, industrial safety, and public health in
affected regions. Furthermore, the findings and methodologies developed in this research
will provide valuable insights and guidelines for future research in the field of
industrial smoke detection and human-in-the-loop machine learning, particularly
in the context of limited annotations and domain adaptation challenges.

\chapter{Related Work} 


\ifpdf \graphicspath{{7/figures/PNG/}{7/figures/PDF/}{7/figures/}} \else
\graphicspath{{7/figures/EPS/}{7/figures/}} \fi

%
\section{Smoke Segmentation Methods}
\label{sec:Smoke Segmentation Methods} Industrial smoke segmentation is a
crucial task in various industries, including power plants, oil refineries, and chemical
processing facilities. Deep learning has been widely adopted for this task due
to its ability to learn complex features from large datasets. Khan et al.~\cite{KHAN2021115125}
propose a deep learning-based approach, \textit{DeepSmoke}, a two-module framework
that combines an \textit{EfficientNet}-based smoke detector with a \textit{DeepLabv3+}
encoder-decoder for semantic segmentation in outdoor (wildfire) environments~\cite{chen2018encoderdecoderatrousseparableconvolution,
tan2020efficientnetrethinkingmodelscaling}. Yan et al.~\cite{yan2023transmissionguidedbayesiangenerativemodel}
propose a transmission-guided Bayesian generative model for smoke segmentation,
which explicitly models uncertainty from limited data and uses a physics-inspired
transmission-guided loss to handle the low-contrast nature of smoke. Furthermore,
Wei et al.~\cite{wei2019f3netfusionfeedbackfocus} introduce a fusion-feedback focus
network (\textit{F3Net}) that integrates multi-scale features and feedback
mechanisms to enhance salient object detection. More recently, Liu et al.~\cite{liu2025smokenetefficientsmokesegmentation}
propose \textit{SmokeNet}, an efficient smoke segmentation model that leverages multi-scale
convolutions and multi-view attention mechanisms to improve segmentation accuracy
in complex industrial environments.

While these methods have shown promising results in smoke segmentation tasks,
they often struggle with domain adaptation and generalization to unseen
environments. This limitation highlights a considerable space for advancement,
particularly in the context of industrial toxic cloud segmentation. The performance
of these advanced methods, however, is intrinsically linked to the availability and
quality of training data.

\section{Dataset Scarcity in Industrial Smoke Emission}
\label{sec:Dataset Scarcity in Industrial Smoke Emission} A fundamental challenge
in industrial monitoring is the paucity of labeled data. For instance, manually annotating
a multi-label (30-class) image at the pixel-level from the \textit{Cityscape}
dataset can take more than 90 minutes~\cite{cordts2016cityscapesdatasetsemanticurban}.
Moreover, it is even more expensive and harder to obtain such stringent
annotations in a specific industrial domain, where experts must precisely mark targets
under complex visual conditions~\cite{schwonberg2023surveyunsuperviseddomainadaptation}.
Currently, the largest publicly available and pixel-wise annotated dataset for smoke
segmentation is the \textit{SMOKE5K} dataset~\cite{yan2023transmissionguidedbayesiangenerativemodel},
which contains around 5,000 synthetic and real-world images with pixel-wise annotations
collected from \textit{SYN70K}~\cite{yuan2018deepsmokesegmentation} and a
wildfire dataset~\cite{Zhou2016}. However, this dataset is limited in its
diversity and does not cover all possible scenarios in industrial environments, which
are often characterized by complex backgrounds, varying lighting conditions, and
different types of smoke emissions. To address the scale issue, Cheng et al.~\cite{cheng_gcce19_smoke100k}
propose the large-scale synthetic smoke dataset \textit{Smoke100k}, which
comprises bounding-box-level annotations. However, Smoke100k does not directly
provide pixel-wise annotations, which are essential for training and evaluating segmentation
models in industrial scenarios. The lack of fine-grained, pixel-level labels in Smoke100k
limits its applicability for industrial smoke segmentation tasks that require precise
boundary delineation. Moreover, Hsu et al.~\cite{hsu2024projectriserecognizingindustrial}
propose the \textit{Project RISE} dataset, a large-scale dataset for recognizing
industrial smoke emissions, yet it only provides video-level coarse labels provided
by local communities and does not include pixel-wise annotations. Therefore, the
scarcity of labeled data remains a significant challenge in industrial smoke segmentation,
underscoring the necessity of developing effective semi- or weakly-supervised methods.
This data-scarce environment necessitates innovative approaches that can learn
from limited or incomplete supervision.

\section{Semi- and Weakly-Supervised Semantic Segmentation}
\label{sec:Semi- and Weakly-Supervised Semantic Segmentation} Semantic
segmentation, a core task in computer vision, requires costly pixel-level
annotations, motivating research into semi-supervised (SSSS) and weakly-supervised
(WSSS) approaches that exploit cheaper forms of supervision. These methodologies
are critical for overcoming the data bottlenecks discussed previously.

\paragraph{Semi-Supervised Methods}
Early SSSS methods blend pseudo-labeling with consistency regularization to propagate
labels from a small annotated set to unlabeled images. Chai et al.~\cite{10916843}
propose a hierarchical generative model for biomedical segmentation,
disentangling semantic mask synthesis under scarce labels. Ye et al.~\cite{ye2025semikankanprovideseffective}
leverage Kolmogorov--Arnold Networks (KANs) for compact yet expressive feature representations
in medical image segmentation. Zhou et al.~\cite{zhou2025crossfrequencycollaborativetrainingnetwork}
incorporate frequency-domain priors and uncertainty-guided mixing to produce high-confidence
pseudo-labels while preserving structural details. These semi-supervised
techniques demonstrate the potential of leveraging unlabeled data to augment
limited annotated datasets.

\paragraph{Image-Level Weakly-Supervised Methods}
\label{sec:Image-Level Weakly-Supervised Methods} Image-level WSSS methods derive
pixel masks from class activation maps (CAMs), which are then refined by
advanced strategies. Ahn and Kwak~\cite{ahn2018learningpixellevelsemanticaffinity}
introduce \textit{AffinityNet}, which learns pixel-level semantic affinities to propagate
coarse CAMs into precise segmentation masks. Lee et al.~\cite{lee2021antiadversariallymanipulatedattributionsweakly}
propose \textit{AdvCAM}, using adversarial manipulation to expand CAM coverage.
Chen et al.~\cite{chen2022classreactivationmapsweaklysupervised} present \textit{ReCAM}
to mitigate false positives and negatives via a reactivation loss. Lin et al.~\cite{lin2023clipefficientsegmentertextdriven}
introduce \textit{CLIP-ES}, leveraging frozen CLIP models to generate
segmentation masks without additional training. Yoon et al.~\cite{Yoon2024}
employ diffusion-guided refinement in \textit{DiG} to denoise and enhance CAM-based
pseudo-labels. Han et al.~\cite{han2024cobracomplementarybranchfusing} fuse CNN and
ViT~\cite{dosovitskiy2021imageworth16x16words} branches in \textit{CoBra}, combining
complementary class and semantic knowledge for state-of-the-art performance. The
progression of these methods shows a clear path toward extracting finer-grained
information from coarse labels.

\paragraph{Bounding-Box-Level Weakly-Supervised Methods}
\label{sec:Bounding-Box-Level Weakly-Supervised Methods} Box-level supervision provides
an intermediate annotation granularity, offering a balance between cost and
precision. Dai et al.~\cite{dai2015boxsupexploitingboundingboxes} introduce
\textit{BoxSup}, which iteratively recovers high-quality masks from bounding
boxes and segmentation feedback. Song et al.~\cite{song2019boxdrivenclasswiseregionmasking}
extend this via class-wise masking and filling-rate guided losses to filter
irrelevant regions. Reiß et al.~\cite{10204652} generalize weak supervision by
learning from diverse annotation types, including bounding boxes, through
decoupled semantic prototypes.

These developments demonstrate a clear trend toward flexible architectures and
training strategies that integrate multiple supervision granularities, from image-level
tags to bounding boxes, thereby reducing annotation cost while maintaining high
segmentation quality. Such approaches are particularly relevant when adapting models
from a data-rich source domain to a data-scarce target domain.

\section{Domain Adaptation for Semantic Segmentation}
\label{sec:Domain Adaptation for Semantic Segmentation} To address the challenge
of limited labeled data in semantic segmentation, domain adaptation has been
widely adopted to improve the performance of semantic segmentation models on
unseen domains. Traditional unsupervised domain adaptation (UDA) methods align feature
or pixel distributions between a labeled source domain and an unlabeled target domain,
reducing the need for costly manual annotation in new domains. This is a crucial
step toward building generalizable models.

Among these UDA approaches, those based on Generative Adversarial Networks (GANs)
have been particularly influential~\cite{goodfellow2014generativeadversarialnetworks}.
GANs enable unsupervised image-to-image translation via cycle-consistent losses.
For example, \textit{CycleGAN}~\cite{zhu2020unpairedimagetoimagetranslationusing}
transforms source images into the target style without paired data, while \textit{CyCADA}~\cite{hoffman2017cycadacycleconsistentadversarialdomain}
further integrates this idea with semantic segmentation by performing both image-
and feature-level adaptation. Likewise, \textit{Adversarial Discriminative
Domain Adaptation (ADDA)}~\cite{tzeng2017adversarialdiscriminativedomainadaptation}
focuses on aligning feature distributions through adversarial training. These generative
and adversarial approaches lay the groundwork for more sophisticated adaptation techniques.

More recently, research has increasingly explored the integration of weak labels
for more targeted and effective adaptation, marking a key transition from traditional
UDA to weakly-supervised domain adaptation (WDA). For example, Paul et al.~\cite{paul2020domainadaptivesemanticsegmentation}
propose a framework that utilizes image-level weak labels in the target domain
to enable category-wise domain alignment, demonstrating that leveraging human-provided
feedback or weak supervision can further enhance model performance in challenging
real-world scenarios. Following this trend, recent works have further refined WDA
techniques. Notably, Wang et al.~\cite{Wang_2019} propose a weakly supervised
adversarial domain adaptation method specifically for semantic segmentation in
urban scenes. Their key contribution lies in employing multi-granularity domain discriminators
to achieve more refined feature alignment at both pixel and image levels, while
leveraging weak supervision to guide the adversarial training and refine pseudo-labels.
More recently, Das et al.~\cite{Das2023} introduce a novel approach for weakly-supervised
domain adaptive semantic segmentation by incorporating a prototypical contrastive
learning framework. The success of these methods suggests that even coarse
annotations can significantly improve adaptation outcomes, a concept that can be
extended to non-expert annotators. This naturally leads to the exploration of
citizen science as a promising approach for acquiring large-scale weak labels in
industrial applications.

\section{Citizen Science in Deep Learning Workflows}
\label{sec:Citizen Science in Deep Learning Workflows} Crowdsourcing is a
powerful tool that enables large amounts of the annotation effort to be
performed by a crowd of citizen scientists. The idea of leveraging non-expert human
contributions to assist in training has gained interest across various domains.
In medical imaging, Spicher et al.~\cite{Spicher2024} developed an integrated
crowdsourcing platform for the segmentation of eye-fundus images. In biology,
Bafti et al.~\cite{BAFTI2021104204} utilized crowdsourcing to annotate microbiological
images of gut parasites and compared the results with annotations by experts.
Pennington et al.~\cite{Pennington2024} conducted a large-scale study via a Zooniverse
project where volunteers were asked to provide point annotations of viruses in cryo-EM
volumes. Furthermore, in the context of industrial toxic cloud segmentation,
\textit{Project RISE}, proposed by Hsu et al.~\cite{hsu2024projectriserecognizingindustrial},
has shown the potential of leveraging citizen scientists to provide coarse
annotations for smoke emissions in industrial environments. Previous studies have
consistently shown that citizen scientists can provide high-quality weak labels that
are comparable to those of experts, while significantly reducing the cost and
time required for manual annotation.

\section{Positioning and Contributions}
\label{sec:Positioning and Contributions}

Built upon the backbone of the \textit{Transmission-guided Bayesian} (TGB)
network~\cite{yan2023transmissionguidedbayesiangenerativemodel} and local
contrastive learning~\cite{chaitanya2021localcontrastivelosspseudolabel}, we propose
\textbf{CEDANet} (Citizen-Engaged Domain-Adaptive Network), a weakly-supervised domain
adaptation framework that explicitly incorporates citizen-provided weak labels to
guide more precise knowledge transfer. Our main contributions are twofold:

\begin{enumerate}
    \item \textbf{Citizen-Science-Informed Pseudo-Label Generation.} We leverage
        coarse, video-level annotations provided by non-expert citizens~\cite{hsu2024projectriserecognizingindustrial}
        to drive a lightweight pseudo-label selection and refinement pipeline,
        reducing reliance on expensive pixel-level expert masks and human effort.

    \item \textbf{Category-Aware Domain Adaptation for Industrial Smoke
        Segmentation.} Inspired by Paul et al.~\cite{paul2020domainadaptivesemanticsegmentation},
        we extend their WDA paradigm to bridge existing smoke datasets and the
        industrial toxic-cloud segmentation domain, enabling fine-grained alignment
        of smoke vs.\ background features under limited annotations. Specifically,
        we go beyond image-level labels by converting inexpensive video-level
        labels into dense, pixel-wise pseudo-labels, thereby injecting spatially
        detailed weak supervision signal into the domain adaptation process.
\end{enumerate}

To our knowledge, no prior work combines category-aware domain adaptation with
citizen-driven pseudo-labeling for industrial smoke segmentation. CEDANet thus pioneers
this integration, addressing both domain shift and annotation scarcity in real-world
monitoring applications. Our source code is publicly available at:
\url{https://github.com/Tao-Yida/CEDANet} to facilitate reproducibility and
further research.

\chapter{Our Methods} 


\ifpdf \graphicspath{{7/figures/PNG/}{7/figures/PDF/}{7/figures/}} \else
\graphicspath{{7/figures/EPS/}{7/figures/}} \fi

This chapter provides a detailed technical exposition of our proposed \textbf{CEDANet}
(Citizen-Engaged Domain-Adaptive Network). As positioned in Section
\ref{sec:Positioning and Contributions}, CEDANet is a human-in-the-loop, weakly-supervised
domain adaptation framework designed specifically for industrial smoke
segmentation. It integrates a robust segmentation backbone with a class-aware adaptation
mechanism, guided by weak labels from citizen scientists.

To provide a clear picture of our methodology, we first present a high-level overview
of the entire framework in Section~\ref{sec:Framework Overview}. This section
explains the multi-stage pipeline and clarifies the roles of its core components,
addressing the main contributions and the overall workflow. Following this
overview, we delve into the specific technical details. Section~\ref{sec:Foundational
Components}
introduces the foundational TGB network and the contrastive loss function we
employed. Section~\ref{sec:Human-in-the-Loop Weakly Supervised Domain Adaptation}
then elaborates on the core innovations of CEDANet: the human-in-the-loop
pipeline for pseudo-label generation and refinement and the class-aware domain
adaptation architecture. Section~\ref{sec:Implementation Details} provides a comprehensive
overview of our experimental setup, including datasets, evaluation metrics, and
training protocols.

\section{Framework Overview}
\label{sec:Framework Overview}

We introduce CEDANet, a novel framework that bridges the crucial gap between
high-level, inexpensive human supervision and the demand for dense, pixel-level
annotations in domain-adaptive industrial smoke segmentation. The core idea is to
\textbf{orchestrate} a powerful pre-trained segmentation model within a workflow
consisting of a human-guided filtering mechanism and a class-aware adversarial
network.

Our contribution is not a new standalone module, but rather a novel \textbf{methodology
for supervision transformation}. We realize this through the synergistic integration
of existing components into a cohesive framework, whose core function is to
convert sparse, video-level classifications into powerful, pixel-wise pseudo-masks.
This process effectively fuses strong labels from the source domain with these refined
pseudo-labels from the target domain, enabling a highly targeted and efficient
adversarial adaptation. The entire workflow can be broken down into two main stages.

\textbf{Stage 1: Supervision Cascade --- From Weak Human Insight to Pseudo-labels.}
The first stage focuses on creating a high-quality labeled dataset for the
target domain. We start with a \textit{Transmission-guided Bayesian (TGB)
network}, pre-trained on a source dataset with rich annotations (e.g., SMOKE5K),
which we denote as $\mathcal{M}_{pretrain}$. This model is used to perform \textit{inference}
on unlabeled target domain videos, generating initial pixel-wise pseudo-labels (segmentation
masks). However, these labels can be noisy. Therefore, we introduce \textit{human
feedback} in the form of binary, video-level labels (i.e., "smoke" or "no smoke")
provided by citizen scientists.

This approach marks a significant innovation compared to prior work in weakly-supervised
domain adaptation~\cite{paul2020domainadaptivesemanticsegmentation}, which
typically relies on image-level weak labels that only indicate the presence or
absence of an object category within an image. In contrast, our method elevates
the role of weak supervision at a data curation stage by converting simple, cost-effective
video-level annotations into pixel-level pseudo-masks. This transformation
provides a much richer and more granular supervisory signal, guiding the model with
spatial information that is absent in traditional image-level labels.

This cascade is used to \textit{select} and \textit{refine} the model-generated pseudo-labels
through a heuristic-based strategy, ensuring that the selected labels are
consistent with human judgment and temporally stable. This process, detailed in
Section \ref{sec:Human-in-the-Loop Pseudo Label Refinement}, yields a refined target
domain dataset with reliable pseudo-masks, ready for the next stage.

\textbf{Stage 2: Class-Aware Domain Adaptation.} The second stage aims to adapt the
model to the specific visual characteristics of the target domain. For this, we employ
a \textbf{Domain Adaptation Framework}, whose core is a feature generator $G_{DA}$
that shares the same backbone architecture as the pre-trained $\mathcal{M}_{pretrain}$.
The training is performed in an adversarial, GAN-like manner, where the
generator $G_{TGB}$ learns to extract \textit{domain-invariant features} that can
"fool" a discriminator. Crucially, to address the unique challenges of smoke segmentation,
we employ two specialized, \textbf{class-aware discriminators} ($D_{smoke}$ and $D
_{bg}$), which learn to distinguish domains for smoke and background features independently.
This separation is vital to prevent negative transfer, where aligning generic background
features (like steam or cloud, which resemble smoke) inadvertently corrupts the
fine-grained foreground features essential for accurate smoke detection. This domain
adaptation mechanism is further explained in Section \ref{sec:Class-Aware Domain
Adaptation Framework}.

Upon completion of this end-to-end training, the generator from the domain
adaptation framework, $G_{DA}$, serves as the final, fine-tuned segmentation model,
capable of producing high-fidelity segmentation masks on new target domain
images.

\section{Foundational Components}
\label{sec:Foundational Components}
\subsection{Transmission-guided Bayesian Network}
\label{sec:Transmission-guided Bayesian Network} Smoke segmentation is
challenging due to the transparent, non-rigid, and amorphous nature of smoke, which
makes it difficult to distinguish boundaries from the background. Moreover, smoke
images are often compromised by sensor noise, particularly affecting
deterministic deep learning approaches with biased datasets.

To address these challenges, we adopt the Transmission-guided Bayesian (TGB) generative
network introduced by Yan et al.
\cite{yan2023transmissionguidedbayesiangenerativemodel} as our backbone model. The
TGB network incorporates a variational auto-encoder (VAE) architecture \cite{kingma2022autoencodingvariationalbayes}
that captures both aleatoric (data) and epistemic (model) uncertainties \cite{hu2022superviseduncertaintyquantificationsegmentation},
making it particularly suitable for handling the inherent ambiguity in smoke
segmentation tasks.

\begin{figure}
    \centering
    \includegraphics[width=0.8\textwidth]{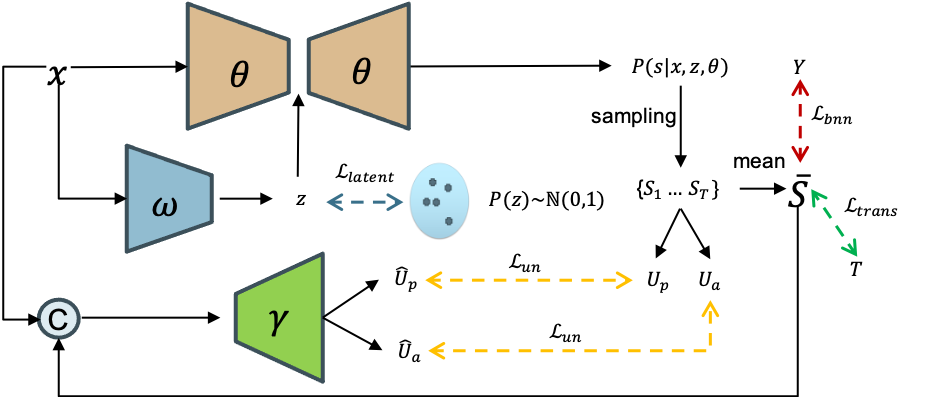}
    \caption{The architecture of the Transmission-guided Bayesian generative
    network, which serves as the backbone of our model \cite{yan2023transmissionguidedbayesiangenerativemodel}.}
    \label{fig:TGB}
\end{figure}

The TGB network comprises three key components that are relevant to our domain adaptation
framework:

\textbf{Bayesian Uncertainty Modeling:} The network uses a VAE structure with
latent variable $z$ to model prediction uncertainties, enabling robust feature learning
under domain shift conditions.

\textbf{Transmission-guided Loss:} Inspired by physics-based dehazing methods, this
component leverages transmission features to distinguish smoke boundaries,
providing valuable physical priors for cross-domain generalization.

\textbf{Uncertainty Calibration:} An entropy-based calibration mechanism prevents
overconfident predictions in ambiguous regions, which is crucial for reliable
pseudo-label generation in our human-in-the-loop framework.

The overall loss function combines these components as:
\begin{equation}
    \mathcal{L}_{gen}\,=\,\mathcal{L}_{ELBO}\,+\,\lambda_{1}\,\mathcal{L}_{trans}
    \,+\,\lambda_{2}\,\mathcal{L}_{c},
\end{equation}
where $\mathcal{L}_{ELBO}$ is the evidence lower bound, $\mathcal{L}_{trans}$ is
the transmission-guided loss, and $\mathcal{L}_{c}$ is the uncertainty calibration
loss. The hyperparameter $\lambda_{1}$ and $\lambda_{2}$ balance the
contributions of each term.

\subsection{Contrastive Loss for Smoke Segmentation}
\label{sec:Contrastive Loss for Smoke Segmentation}

To enhance feature learning in our weakly-supervised setting, we employ the
local contrastive loss proposed by Chaitanya et al.~\cite{chaitanya2021localcontrastivelosspseudolabel}.
Unlike traditional contrastive methods relying on data augmentation or spatial
proximity~\cite{chen2020simpleframeworkcontrastivelearning,he2020momentumcontrastunsupervisedvisual},
this method defines pixel-level semantic similarity based on pseudo-labels,
enabling semantic-aware feature representation learning.

Specifically, the contrastive loss encourages pixels belonging to the same class
to form compact feature clusters and enforces separability across classes. We
adapt this strategy to our binary segmentation task (smoke vs. background).
Given two image-label pairs $(x,y)$ and $(x',y')$, where $x$, $x'$ are input
images and $y$, $y'$ represent pixel-wise pseudo-labels, the adapted pixel-wise
contrastive loss is defined as:
\begin{equation}
    \mathcal{L}_{\text{cont}}((x, y), (x', y'))\,=\,\frac{1}{|S_{\text{smoke}}(x)|}
    \sum_{i \in S_{\text{smoke}}(x)}L_{i,\text{smoke}}\left([z(x)]_{i}, \bar{z}_{\text{smoke}}
    (x')\right),
\end{equation}
where $S_{\text{smoke}}(x)$ denotes the set of smoke-class pixel indices in image
$x$, $[ z(x)]_{i}$ is the feature vector of pixel $i$, and $\bar{z}_{\text{smoke}}
(x')$ represents the average feature vector (class center) of smoke pixels in reference
image $x'$, computed as:
\begin{equation}
    \bar{z}_{\text{smoke}}(x')\,=\,\frac{1}{|S_{\text{smoke}}(x')|}\sum_{i \in
    S_{\text{smoke}}(x')}[z(x')]_{i}.
\end{equation}
The per-pixel loss function $L_{i,\text{smoke}}$ utilizes cosine similarity between
the pixel feature vector and the smoke class center:
\begin{equation}
    L_{i,\text{smoke}}\left([z(x)]_{i}, \bar{z}_{\text{smoke}}(x')\right)\,=\,-\log
    \frac{e^{\text{sim}([z(x)]_{i}, \bar{z}_{\text{smoke}}(x'))/\tau}}{e^{\text{sim}([z(x)]_{i},
    \bar{z}_{\text{smoke}}(x'))/\tau}+ e^{\text{sim}([z(x)]_{i}, \bar{z}_{\text{bg}}(x'))/\tau}}
    ,
\end{equation}
where
$\text{sim}(\mathbf{a}, \mathbf{b})\,=\,\frac{\mathbf{a}^{T}\mathbf{b}}{||\mathbf{a}||\cdot
||\mathbf{b}||}$
denotes the cosine similarity and $\tau$ is a temperature parameter controlling
the sharpness of the feature distribution.

By applying this loss, the model learns a representation where the smoke pixels are
closely grouped while effectively separated from the background pixels, thus
improving the semantic discrimination capability for smoke segmentation. We borrow
the implementation of the local contrastive loss from~\cite{MultiX-Amsterdam-ijmond-camera-ai}.

\section{Human-in-the-Loop Weakly Supervised Domain Adaptation}
\label{sec:Human-in-the-Loop Weakly Supervised Domain Adaptation}

In this section, we present our human-in-the-loop weakly supervised domain adaptation
framework, which is designed to leverage citizen-provided weak labels in the
target domain alongside model-generated pseudo-labels in a class-aware domain
adaptation setting. The key idea of this framework is to keep the domain adaptation
process class-aware, meaning that the model is trained to adapt to the target
domain while preserving the class-wise structure of the feature space. Then, the
model can effectively utilize the weak labels provided by citizens to refine the
pseudo-labels generated by a pre-trained model, which further improves the performance
of the DA model.

\subsection{Human-in-the-Loop Pseudo Label Refinement}
\label{sec:Human-in-the-Loop Pseudo Label Refinement}

In real-world smoke segmentation tasks, where it is hard and expensive to acquire
expert-level, pixel-wise annotations, citizen science has emerged as a promising
and scalable solution to address these challenges by mobilizing the collective efforts
of non-experts to provide weak labels, such as high-level binary judgments, scribbles
and bounding boxes \cite{paul2020domainadaptivesemanticsegmentation, hsu2024projectriserecognizingindustrial}.
In our research, we leverage \emph{citizen feedback}, which is in the form of video-level
binary labels (i.e., smoke vs. no smoke) provided by local communities from the
IJmond region, to select and refine the pseudo-labels generated by pretrained models
\cite{breathecam2024}. Specifically, the volunteers have provided coarse labels for
each video, indicating if it has smoke or not, which serves as a weak supervision
signal for our model. Likewise, all the videos have also been inspected by professional
researchers, giving each an admin label. In this scenario, the citizens serve as
a \textbf{Weak Oracle} that supervises and constrains the generation of pseudo-labels,
guiding the model's prediction to keep consistent with human judgment and
reducing possible false positives \cite{Kosmala2016}.

\begin{figure}[htb]
    \centering
    \includegraphics[width=0.9\textwidth]{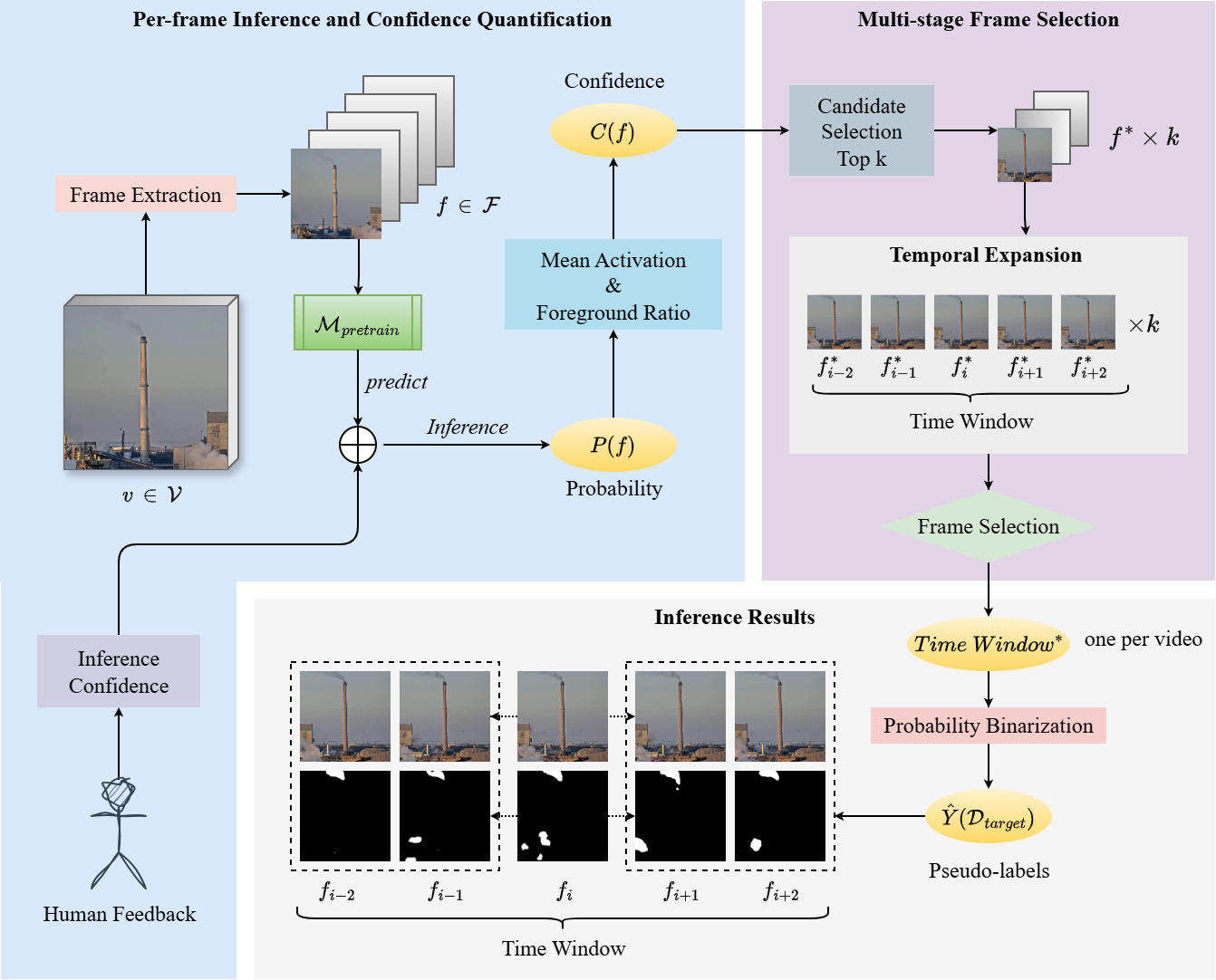}
    \caption[The Frame Selection and Pseudo-Label Generation Pipeline]{Overview
    of the frame selection and pseudo-label generation pipeline. The process begins
    by extracting frames $f$ from a video $v$. A pretrained model
    $\mathcal{M}_{pretrain}$ generates an initial probability map $P(f)$, from which
    a confidence score $C(f)$ is derived. Top-$k$ candidate frames $f^{*}$ are
    selected based on confidence and expanded into a temporal window. After a final
    selection step, probability binarization is applied to the frames in the
    optimal time window, producing pseudo-labels $\hat{Y}(\mathcal{D}_{target})$
    for the selected frames.}
    \label{fig:human-feedback-pseudo-labeling}
\end{figure}

\paragraph{Video Label}
In our framework, each video is assigned a binary label and a corresponding \textit{inference
confidence} score derived from volunteer feedback, as detailed in Table~\ref{tab:label_confidence}.
It should be noted that even expert annotations may contain errors; therefore,
we avoid using extreme confidence values (i.e. 0 or 1) to enhance the model robustness.

\begin{table}[htb]
    \centering

    \begin{tabular}{cccc}
        \hline
        Label Value & Label Meaning          & Expected Smoke & Inference Confidence \\
        \hline
        47          & Gold Standard Positive & True           & 0.9                  \\
        32          & Gold Standard Negative & False          & Skip                 \\
        23          & Strong Positive        & True           & 0.8                  \\
        16          & Strong Negative        & False          & Skip                 \\
        19          & Weak Positive          & True           & 0.7                  \\
        20          & Weak Negative          & False          & Skip                 \\
        5           & Maybe Positive         & True           & 0.65                 \\
        4           & Maybe Negative         & False          & Skip                 \\
        3           & Disagreement           & Unknown        & No constraint        \\
        -1          & No Data                & Unknown        & No constraint        \\
        \hline
    \end{tabular}
    \caption{Video-level Weak Labels and Corresponding Inference Confidence.}
    \label{tab:label_confidence}
\end{table}

Furthermore, the \textit{golden standard} labels are exclusively provided by
researchers, whereas other label categories are shared. If a video is marked as
\textit{Disagreement} or \textit{No Data}, it is treated as having \textit{No
constraint}, meaning the final frame-level pseudo-label is determined solely by the
model's prediction. This mapping ensures a flexible balance between human
supervision and model autonomy.

In our implementation, when fusing the model prediction and the human-provided video-level
constraint, we use a weighted sum
$P_{fused}= \lambda P_{model}+ (1-\lambda_{conf}) P_{constraint}$, where $\lambda
_{conf}$ is set adaptively: for positive (smoke) labels, $\lambda_{conf}=0.8$ (model-dominant),
and for negative (no smoke) labels, $\lambda_{conf}=0.4$ (constraint-dominant).
This ensures that negative samples are more strictly constrained by human
annotation, while positive samples allow greater flexibility for model prediction.

\paragraph{Frame-level Pseudo-label Generation and Selection}
In our framework, we do not train a separate image classification model for frame
selection. Instead, we employ a \textbf{heuristic-based multi-stage selection
strategy} that leverages the inference results from the pre-trained segmentation
model. This strategy is designed to automatically identify the most representative
and temporally stable frames for pseudo-label generation. The core idea is that
an ideal frame should not only contain clear smoke features but also maintain
high content consistency with its neighboring ones~\cite{Miksik2013}. The selection
process is divided into the following key stages:

\begin{enumerate}
    \item \textbf{Per-Frame Inference and Confidence Quantification.} The pre-trained
        smoke segmentation model is first applied to all sampled frames of a
        video. For each frame $f$, the model outputs a pixel-wise smoke
        probability map $P(f) \in [0, 1]^{H \times W}$. We then quantify the model's
        confidence for each frame using a composite score that combines two metrics:
        \begin{itemize}
            \item \textit{Mean Activation}: The average pixel value of the probability
                map, reflecting the overall likelihood of smoke presence.

            \item \textit{Foreground Ratio}: The proportion of pixels in the probability
                map exceeding a high-confidence threshold $\tau$ (e.g.,
                $\tau=0.6$), measuring the extent of the high-certainty smoke
                region.
        \end{itemize}
        The confidence score $C(f)$ for a frame $f$ is calculated as:
        \begin{equation}
            C(f) = \alpha \cdot \mathbb{E}[P(f)] + (1 - \alpha) \cdot \frac{\#\{p
            \in P(f) \mid p > \tau\}}{H \times W},
        \end{equation}
        where $\mathbb{E}[P(f)]$ is the mean activation, $H \times W$ is the
        frame resolution, and $\alpha$ is a weighting factor (set to 0.85 in our
        implementation). This design favors predictions that are both confident
        and spatially significant.

    \item \textbf{Identification of Candidate Center Frames.} After computing
        the confidence score for all frames, we sort them in descending order.
        Instead of selecting only the top-scoring frame, we identify the top-$k$
        frames (e.g., $k=3$) as \textit{candidate center frames} $f^{*}$. This approach
        enhances the robustness of our selection by considering multiple high-quality
        candidates.

    \item \textbf{Temporal Coherence Assessment.} For each candidate center
        frame, we form a temporal group consisting of the frame itself and its $\pm
        n$ temporal neighbors (e.g., $n=2$). We then evaluate the \textit{intra-group
        similarity} to measure the stability of the model's predictions within this
        short time window. The similarity between any two frames $u$ and $v$ in
        the group is a composite metric:
        \begin{equation}
            \mathrm{sim}(u, v) = \beta \cdot \mathrm{IoU}(\hat{y}(u), \hat{y}(v))
            + (1 - \beta) \cdot \mathrm{corr}(P(u), P(v)),
        \end{equation}
        where $\hat{y}$ is the binarized mask obtained by thresholding the probability
        map $P$ at $\tau$, $\mathrm{IoU}$ is the Intersection over Union, $\mathrm{corr}$
        is the Pearson correlation coefficient between the raw probability maps,
        and $\beta$ is a weighting factor (set to 0.7). We then compute the average
        similarity $\overline{\mathrm{sim}}$ for the entire group.

    \item \textbf{Final Group Selection.} Finally, we calculate a composite score
        for each candidate group, combining the confidence of its center frame
        with the group's temporal coherence:
        \begin{equation}
            \mathrm{Score}_{\text{group}}= \gamma \cdot C(f_{\text{center}}) + (1
            - \gamma) \cdot \overline{\mathrm{sim}}_{\text{group}},
        \end{equation}
        where $\gamma$ is a weighting factor (set to 0.8). The temporal group
        with the highest composite score is selected, and all frames within this
        group are used for generating pseudo-labels. This ensures that our
        selected frames are not only highly confident but also temporally
        consistent, providing high-quality and low-noise supervision for subsequent
        training. If no valid group can be formed (e.g., for very short videos or
        the model cannot extract any salient smoke area), the system defaults to
        selecting the single frame with the highest confidence score and its immediate
        neighbors.
\end{enumerate}

By following this procedure, we ensure that the pseudo-labels are not only consistent
with the human-provided video-level weak labels but also robust against noise in
the model's predictions. The final pseudo-labels are then used to train the
domain adaptation model, allowing it to learn from both the weak human labels and
the model-generated pseudo-labels, thus improving its performance on the smoke
segmentation task.

\subsection{Class-Aware Domain Adaptation Framework}
\label{sec:Class-Aware Domain Adaptation Framework} Our approach is inspired by
Paul et al.~\cite{paul2020domainadaptivesemanticsegmentation}, whose core
principle dictates that feature distributions of different classes should be
aligned independently, rather than performing a global, class-agnostic alignment.
For the smoke detection task, this implies that the model must learn to
distinguish between smoke and background features from both the source domain (e.g.,
synthetic data) and the target domain (e.g., real-world low-quality surveillance
footage), and subsequently minimize the distance between their respective feature
space representations. This class-aware strategy effectively prevents negative
transfer, where, for instance, the alignment of background features could interfere
with the effective learning of smoke features. This is particularly crucial for
smoke, which often presents as a mutable and subtly-featured object.

\paragraph{Core Components}
\label{sec:Core Components} Different from those which apply adversarial domain
adaptation at the feature level (e.g. DANN~\cite{ganin2015unsuperviseddomainadaptationbackpropagation},
ADDA~\cite{tzeng2017adversarialdiscriminativedomainadaptation}), our framework is
based on a Bayesian generative segmentation network, and incorporates an
adversarial domain adaptation module with class-aware discrimination. Unlike typical
GANs, our method does not generate images, but instead aligns feature distributions
between domains via class-aware domain discriminators and a gradient reversal layer.
As shown in Figure~\ref{fig:WDA-framework}, it consists of three primary components:

\begin{figure}[htb]
    \centering
    \includegraphics[width=1.0\textwidth]{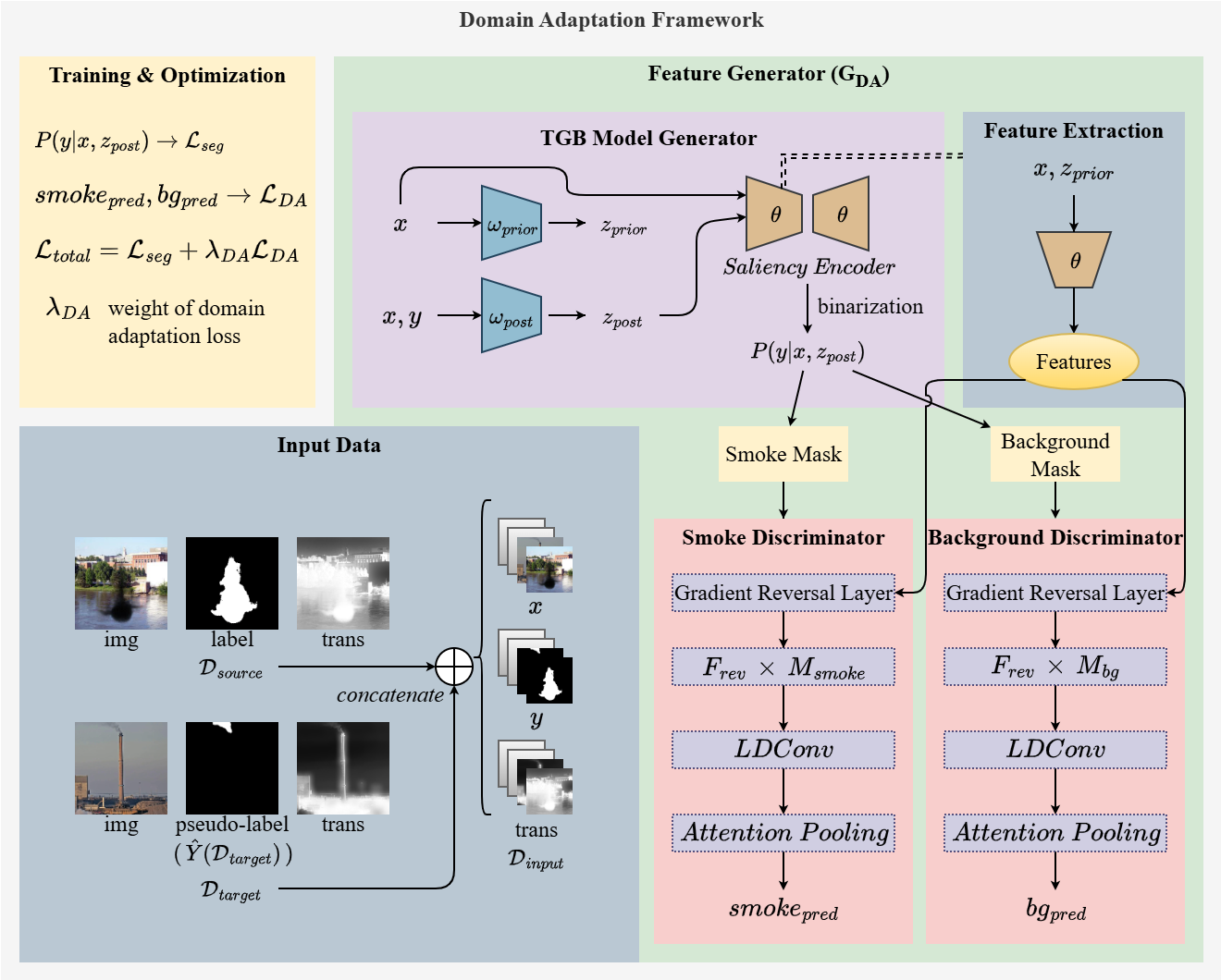}
    \caption{Overview of our weakly supervised domain adaptation framework. The model
    consists of a feature generator, a gradient reversal layer, and two category-aware
    domain discriminators. The feature generator is based on the Transmission-guided
    Bayesian (TGB) network
    \cite{yan2023transmissionguidedbayesiangenerativemodel}.}
    \label{fig:WDA-framework}
\end{figure}

\begin{itemize}
    \item \textbf{Feature Generator (Generator)}: The feature generator $G_{DA}$
        incorporates the original TGB network \cite{yan2023transmissionguidedbayesiangenerativemodel}
        as the backbone, which is responsible not only for accurately segmenting
        smoke regions but also for producing \textbf{domain-invariant features} that
        can deceive the discriminator.

    \item \textbf{Class-Aware Domain Discriminator (Discriminator)}: Unlike conventional
        methods, we employ two separate discriminators for the \textit{smoke} and
        \textit{background} classes, denoted as $D_{smoke}$ and $D_{bg}$,
        respectively. Each discriminator is trained, solely responsible for
        classifying whether the features of its assigned category originate from
        the source or target domain, while not interfering with the other one.
        This design ensures the feature distributions of the two classes are
        aligned independently, thus preventing negative transfer.

    \item \textbf{Gradient Reversal Layer (GRL)}: Proposed by Ganin \cite{ganin2015unsuperviseddomainadaptationbackpropagation},
        the Gradient Reversal Layer is a crucial module in our model, bridging the
        generator and the discriminators. During back propagation, it negates the
        gradients flowing from the discriminators, enabling an adversarial
        optimization process that compels the generator to learn domain-invariant
        features.
\end{itemize}

In the following, we detail each component's function, implementation, and
contribution to the overall adaptation process.
\paragraph{Feature Generator}
\label{sec:Feature Generator} Our feature generator, denoted as $G_{DA}$, is
tasked with two complementary objectives: (i) producing accurate smoke segmentation
masks, and (ii) extracting domain-invariant feature representations.

We build our generator $G_{DA}$ by wrapping the base generator from the VAE-architectured
TGB network, which facilitates multi-task learning by enabling shared feature representations
for both domain discrimination and saliency segmentation tasks. The input image
$x\in\mathbb{R}^{3\times H\times W}$ is processed through its VAE encoder to
predict the distribution parameters $\mu(x)$ and $\text{logvar}(x)$, which represent
the mean and log-variance of the latent variable $z$. Then we sample the latent
representations $z$ via the reparameterization trick:
\[
    z = \mu(x) + \exp{(\,\frac{1}{2}\,\text{logvar}(x)\,)}\odot\epsilon,\quad \epsilon
    \sim \mathcal{N}(0,I)
\]
The sampled latent vector $z$ is spatially tiled and concatenated with the input
image, forming enriched feature maps that are processed through the intermediate
layers of the ResNet~\cite{he2015deepresiduallearningimage} backbone. This design
maintains a balance between spatial resolution and semantic richness, ensuring
high segmentation performance while enriching the feature space, for robust domain-adaptive
learning.

Crucially, $G_{DA}$ is explicitly trained to learn domain-invariant features, i.e.,
feature representations whose distributions are similar between the source and target
domains. This is achieved via adversarial learning against class-aware
discriminators. During training, the feature generator is not only updated to
minimize segmentation error on labeled source images, but also receives adversarial
gradient signals from the discriminators. These adversarial signals encourage
the generator to produce features for target domain samples that are
indistinguishable from those of the source domain, thereby closing the domain gap
at the feature level~\cite{paul2020domainadaptivesemanticsegmentation}. By
balancing the segmentation loss and the adversarial loss, the generator learns representations
that are effective for segmentation and agnostic to domain origin. Meanwhile, a VAE
regularization loss (Kullback-Leibler divergence with linear annealing) is included
to prevent posterior collapse by aligning the latent space with the standard normal
prior, further stabilizing the representation learning process~\cite{yan2023transmissionguidedbayesiangenerativemodel}.

\paragraph{Class-Aware Domain Discriminator}
\label{sec:Class-Aware Domain Discriminator} To effectively address domain shift
in a class-aware manner, we employ \emph{\textbf{two}} class-specific domain discriminators
for the smoke and background classes. Formally, our domain discriminator $D$ receives
an input feature map $F\in\mathbb{R}^{C\times H\times W}$ extracted from the third
layer of the saliency feature generator. A class-wise mask filter, denoted as $M\in
\mathbb{R}^{1\times H\times W}$, is then applied to isolate class-specific
regions (smoke or background) from the feature map.

Following the framework of Paul et al. \cite{paul2020domainadaptivesemanticsegmentation},
we introduce an \textit{attention-guided pooling} layer to further enhance the class-specific
aggregation of spatial features within each domain discriminator. Given a masked
feature map $F \in \mathbb{R}^{C \times H \times W}$ and a corresponding class
attention map $A \in \mathbb{R}^{1 \times H \times W}$ (the predicted probability
map for the target class), it computes a weighted average over the spatial domain,
where higher weights are assigned to more class-relevant regions. Formally, the
pooled feature $f_{c}\in \mathbb{R}^{C}$ is obtained as follows:
\begin{equation}
    f_{c}\,=\,\frac{\sum_{i, j}A_{i, j}\cdot F_{c, i, j}}{\sum_{i, j}A_{i, j}},
\end{equation}
where $c$ indexes the feature channels.

Additionally, we introduce the \textit{Linear Deformable Convolution (LDConv)}
layer, which replaces standard convolution layers in the discriminators, thereby
enhancing the discriminative ability and flexibility in capturing spatial transformations
between domains~\cite{Zhang_2024}. LDConv explicitly learns spatial offsets for
sampling points, allowing the receptive field to dynamically adapt to spatial variations.
This adaptability enables the discriminator to better handle local geometric
misalignments and subtle appearance changes that often exist between the source
and target domains.

In practice, the domain discriminator processes the masked feature map using a sequence
of LDConv layers, followed by a \textit{SiLU} activation function and global pooling
to obtain a domain prediction score. By integrating LDConv, the discriminator can
more effectively focus on relevant spatial structures and context within each class,
leading to improved robustness against complex domain shifts.

As a result, our class-aware domain discriminators equipped with LDConv not only
improve domain discrimination accuracy but also provide more precise and fine-grained
supervision for the generator, which is crucial for effective adaptation in
challenging semantic segmentation scenarios such as industrial smoke detection.

\paragraph{Gradient Reversal Layer (GRL)}
\label{sec:Gradient Reversal Layer (GRL)} The Gradient Reversal Layer is the vital
component that enables the adversarial training between a feature generator $G_{DA}$
and a discriminator $D$ to be implemented seamlessly. Proposed by Ganin et al.~\cite{ganin2015unsuperviseddomainadaptationbackpropagation}
in the context of domain adversarial training, it serves as an operator
$\mathcal{R}$ between $G_{DA}$ and $D$ . During the forward pass, the GRL $\mathcal{R}$
acts as the identity function $\mathcal{R}(F) = F$, passing the features to the
discriminator unchanged. However, different from original GANs, during the backward
propagation, $\mathcal{R}$ negates the gradients back to $G_{DA}$. That is to say,
if the discriminator produces a gradient $\frac{\partial L_{D}}{\partial F}$, the
GRL will pass $\mathcal{R}\left(\frac{\partial L_{D}}{\partial F}\right) = -\lambda_{grl}\frac{\partial
L_{D}}{\partial F}$ back to the feature generator, which \textit{reverses the
direction} of the gradient flow, causing the generator to learn domain-invariant
features that can fool the discriminator and thus facilitating robust domain
alignment.

\paragraph{Training Procedure}
\label{sec:Training Procedure} The training process of our domain adaptation framework
integrates the base TGB generator with category-aware discriminators through a Gradient
Reversal Layer (GRL), which specifically, alternates between optimizing segmentation
accuracy on source domain data and aligning feature distributions between the source
and target domains via adversarial learning.

Given source domain dataset $\mathcal{D}_{S}= \{(x_{s}^{i}, y_{s}^{i})\}^{N_{S}}$
and target domain dataset $\mathcal{D}_{T}= \{(x_{t}^{j}, \hat{y}_{t}^{j})\}^{N_{T}}$,
the objective of this domain adaptation module is to minimize the combined loss
function:
\begin{equation}
    \mathcal{L}_{\text{DA}}\,=\,\frac{1}{4}\left(L^{s}_{\mathcal{D}_{smoke}}\,+\,
    L^{s}_{\mathcal{D}_{bg}}\,+\,L^{t}_{\mathcal{D}_{smoke}}\,+\,L^{t}_{\mathcal{D}_{bg}}
    \right),
\end{equation}
where $L^{d}_{\mathcal{D}_c}\,=\,-\mathbb{E}_{(x,y)\in\mathcal{D}_c}\left[\log D_{c}
( M_{c}\odot F(x))\right]$ is the domain classification loss for class $c$ of the
corresponding domain $d$, and $\mathbb{E}$ is the standard cross-entropy loss. Here,
$\odot$ denotes element-wise multiplication (Hadamard product). The total loss of
our model is:
\begin{equation}
    \mathcal{L}_{\text{total}}\,=\,\mathcal{L}_{\text{gen}}\,+\,\lambda_{\mathrm{cont}}
    \,\mathcal{L}_{\mathrm{cont}}\,+\,\lambda_{DA}\,\mathcal{L}_{\text{DA}}.
\end{equation}

The complete training loop is illustrated in Algorithm~\ref{alg:domain_adaptation_training}.

\begin{algorithm}
    [htb]
    \caption{Domain-Adaptive Training Procedure.}
    \label{alg:domain_adaptation_training}
    \begin{algorithmic}
        [1] \Require Source domain set $\mathcal{D}_{S}$, target domain set
        $\mathcal{D}_{T}$, generator $G_{DA}$, domain discriminators
        $D_{\mathrm{smoke}}, D_{\mathrm{bg}}$, GRL weight parameter $\lambda_{\mathrm{GRL}}$,
        loss weights $\lambda_{\mathrm{DA}}, \lambda_{P}$ \For{epoch = $1 \to N_{\mathrm{epochs}}$}
        \For{each batch} \State Sample
        $(x_{\mathrm{domain}}, y_{\mathrm{domain}})$ from $\mathcal{D}_{S}$ or
        $\mathcal{D}_{T}$ \State
        $\text{output}_{\mathrm{domain}}\gets G(x_{\mathrm{domain}}, y_{\mathrm{domain}}
        ; \lambda_{\mathrm{GRL}})$
        \State
        $F_{\mathrm{domain}}= \mathrm{ExtractFeature}(x_{\mathrm{domain}})$
        \State
        $M_{c}= \mathbb{I}(\sigma(\text{output}_{\mathrm{domain}}) > 0.5)$ for
        $c \in \{\mathrm{smoke}, \mathrm{bg}\}$ \State
        $d^{\mathrm{domain}}_{c}= D_{c}(F_{\mathrm{domain}}, M_{c})$ for
        $c \in \{\mathrm{smoke}, \mathrm{bg}\}$ \State Compute
        $\mathcal{L}^{\mathrm{domain}}_{\mathrm{gen}}= \mathcal{L}_{\mathrm{gen}}
        (\text{output}_{\mathrm{domain}}, y_{\mathrm{domain}})$
        \EndFor \State
        $\mathcal{L}_{\mathrm{seg}}= \mathcal{L}^{S}_{\mathrm{gen}}+ \lambda_{P}\mathcal{L}
        ^{T}_{\mathrm{gen}}$
        \State
        $\mathcal{L}_{\mathrm{DA}}= \frac{1}{4}\sum\limits_{c \in \{\mathrm{smoke},
        \mathrm{bg}\}}[\mathrm{CE}(d^{S}_{c}, 0) + \mathrm{CE}(d^{T}_{c}, 1)]$
        \State
        $\mathcal{L}_{\mathrm{total}}= \mathcal{L}_{\mathrm{seg}}+ \lambda_{\mathrm{DA}}
        \mathcal{L}_{\mathrm{DA}}$
        \State Update all model parameters by minimizing
        $\mathcal{L}_{\mathrm{total}}$ \EndFor
    \end{algorithmic}
\end{algorithm}

\section{Implementation Details}
\label{sec:Implementation Details} To the best of our knowledge, we pioneer the research
direction of systematically applying domain adaptation techniques for industrial
smoke segmentation using synthetic smoke datasets and citizen-engaged pseudo-labeling.
We establish a comprehensive set of experiments to evaluate our method and
introduce a \emph{benchmark} by comparing multiple models on our industrial smoke
data.

\subsection{Dataset}
\label{sec:Dataset}
\paragraph{Source Domain Dataset}
The source domain dataset for our research is the \textbf{SMOKE5K} dataset proposed
by Yan et al. \cite{yan2023transmissionguidedbayesiangenerativemodel}. The fully
annotated training set SMOKE5K contains 5,000 images with pixel-wise annotations,
where 4,000 images are synthetic smoke images selected from the \textbf{SYN70} dataset
\cite{yuan2018deepsmokesegmentation}, 960 real-world wildfire smoke images
selected from \cite{Zhou2016}, and 40 real-world smoke images selected from the Internet.

\begin{figure}[htb]
    \centering
    \begin{minipage}[b]{0.48\textwidth}
        \centering
        \includegraphics[width=\textwidth]{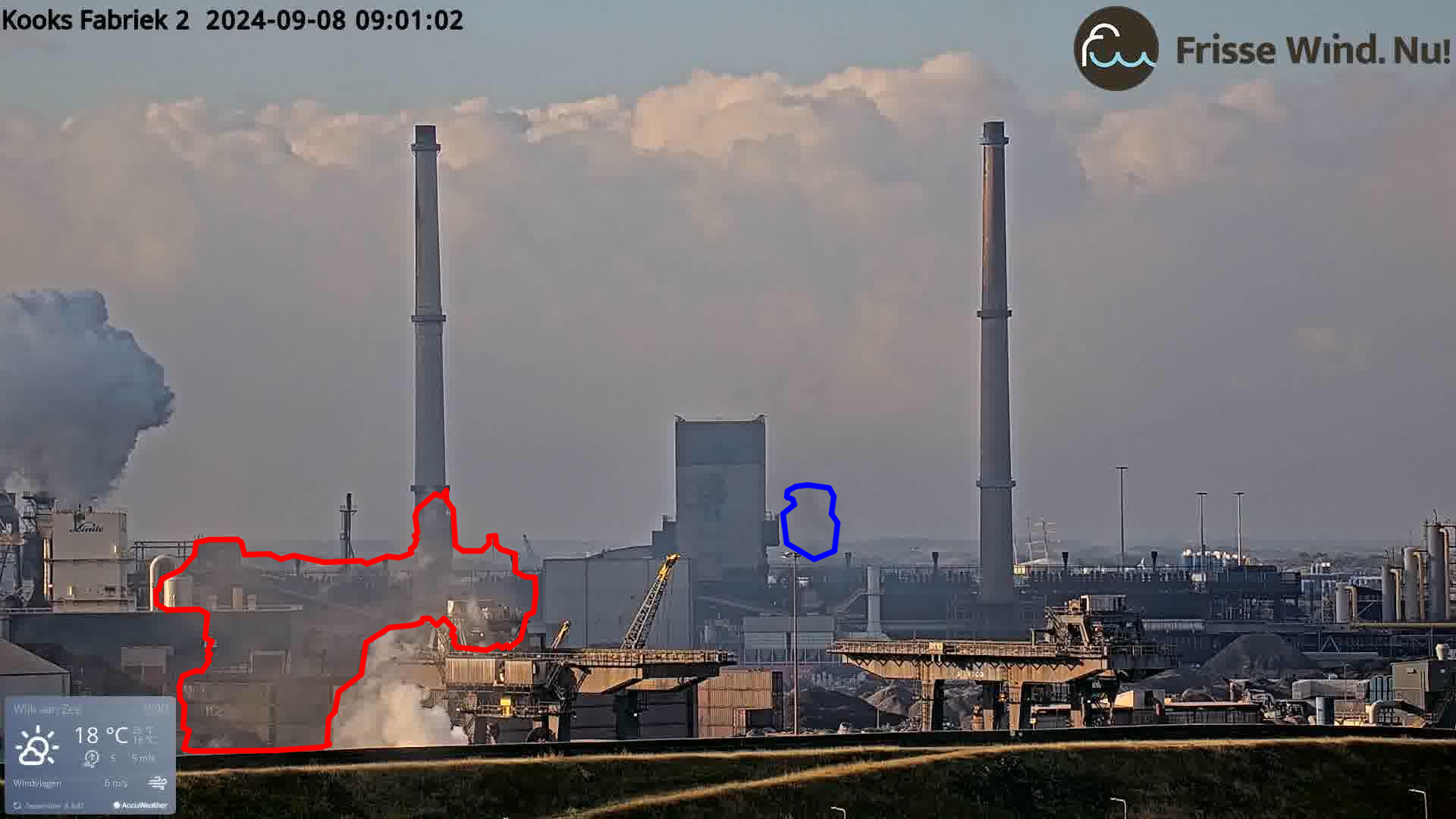}
        {a} \label{fig:sub1}
    \end{minipage}
    \hfill
    \begin{minipage}[b]{0.48\textwidth}
        \centering
        \includegraphics[width=\textwidth]{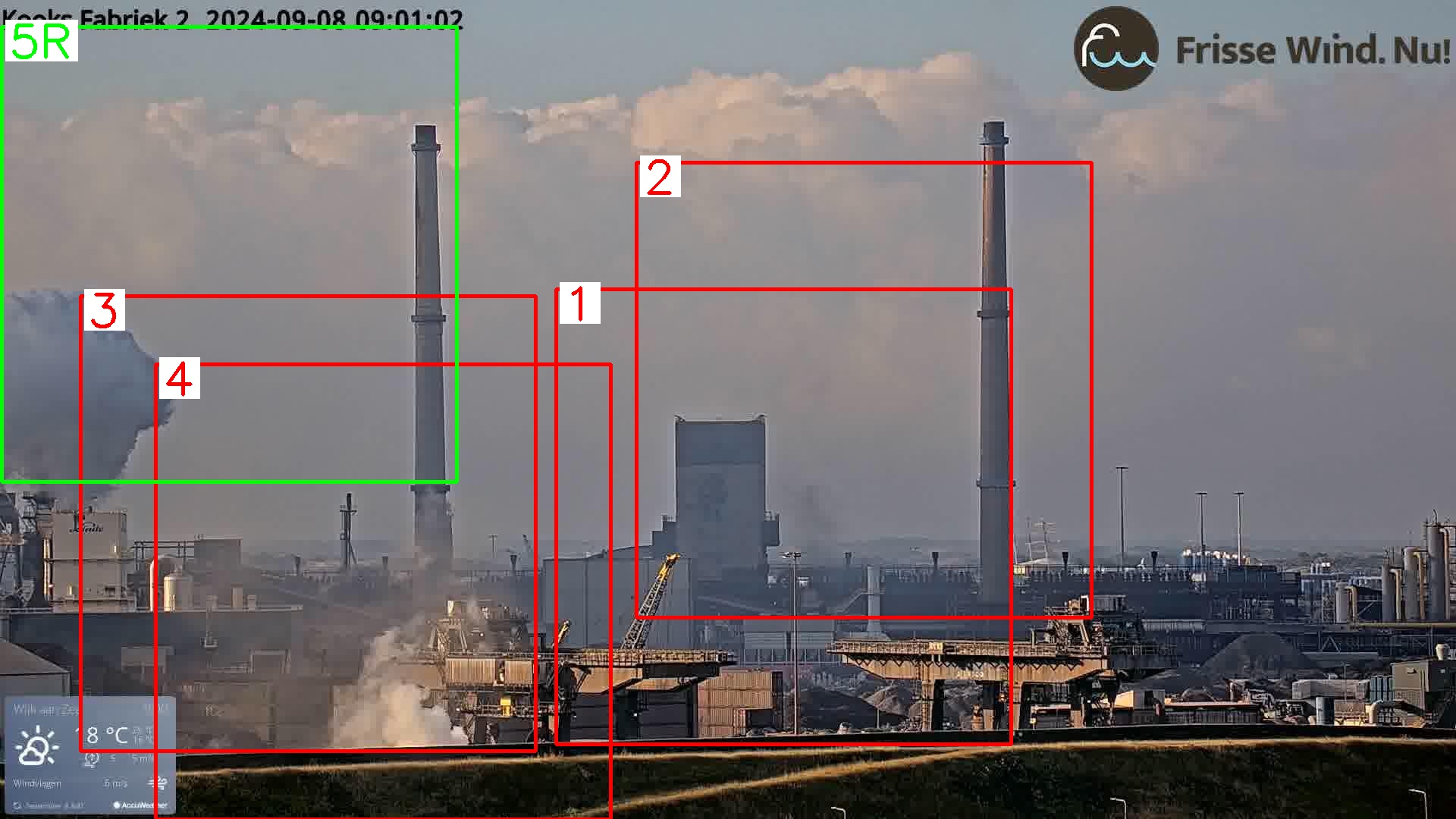}
        {b} \label{fig:sub2}
    \end{minipage}
    \caption{Sample from the target domain IJmond900 dataset. (a) Original image
    with pixel-level ground-truth masks: high-opacity smoke regions in red and
    low-opacity regions in blue. (b) Multi-patch cropping strategy: for each
    annotated smoke area we extract two patches centered with random offsets (red
    boxes), and one additional patch randomly sampled across the image (green
    box).}
    \label{fig:image_cropping_example}
\end{figure}

\paragraph{Target Domain Dataset}
Our target domain dataset comprises two parts: the \textbf{IJmond Camera} video
dataset consisting of 878 one-second videos for pseudo-labeling and the \textbf{IJmond900}
dataset consisting of 900 fully annotated images. These two datasets were
collected from multiple industrial surveillance cameras installed in the IJmond
factory region of the Netherlands \cite{breathecam2024, ijmondcam2024}. Generally,
we use the IJmond Camera dataset for pseudo-labeling and further domain-adapted
model training, and we also use 100 images from IJmond900 as the gold standard for
training, while the remaining 800 images are used for testing and benchmarking
the performance of our proposed method. We also cropped all the images from the IJmond900
testing set into $600 \times 600$ patches to ensure consistency between models's
training and predicting architecture, as illustrated in Fig.~\ref{fig:image_cropping_example}.

\subsection{Evaluation Metrics}
\label{sec:Evaluation Metrics} To comprehensively assess smoke detection
performance, we employ a two-tiered evaluation strategy. For judging the model's
overall performance, we use standard binary segmentation metrics: \textbf{Recall}
is vital for minimizing missed detections, the \textbf{$F_{1}$-Score} offers a balanced
measure of precision and recall, the \textbf{Intersection over Union for the
smoke class ($\mathrm{IoU_{smoke}}$)} evaluates the spatial overlap accuracy specifically
for the foreground, and the \textbf{mean Intersection over Union (mIoU)} evaluates
the overlap accuracy across both smoke and background classes. Additionally, \textbf{mean
Mean Squared Error (mMSE)} quantifies pixel-level intensity accuracy within smoke
regions. This set of metrics provides a robust assessment of segmentation quality,
especially for amorphous smoke shapes.

For a more nuanced analysis of the model's ability to handle smoke's
translucency, we conduct an opacity-specific evaluation against ground-truth masks
with distinct high and low opacity levels. In this context, the metrics of \textbf{Recall},
\textbf{$F_{1}$-Score}, and \textbf{$\mathrm{IoU_{smoke}}$} are extended to
evaluate performance on high- and low-opacity smoke regions separately. This
allows us to quantify the model's effectiveness across the opacity spectrum. Together,
these metrics provide a holistic framework for evaluating both general
segmentation quality and the specific challenges of modeling smoke's varying
opacity.

Additional metrics information can be found in appendix \ref{Appendix:2}.

\subsection{Training Protocol and Experiment Settings}
\label{sec:Training Protocol and Experiment Settings}
\paragraph{Training Details}
We train our model using PyTorch with a maximum of 100 epochs. Each image is
rescaled to $352 \times 352$. Following the methodology proposed by \cite{yan2023transmissionguidedbayesiangenerativemodel},
we set the dimension of the latent space $z$ as 8. The learning rate is set to $5
\times 10^{-5}$, and we use the Adam optimizer and decrease the learning rate
using the ReduceLROnPlateau scheduler with a decay factor of 0.8 and patience
of 12 epochs. Gradient clipping is applied with a threshold of 0.5 to ensure training
stability. For domain adaptation training, we set the domain adaptation loss weight to 0.1. The gradient reversal layer strength $\lambda_{grl}$ is gradually
increased from 0 to 1.0 during training. We use feature channels of 64 dimensions
and apply early stopping with a patience of 15 epochs based on training loss to prevent overfitting. L2 regularization with weight $1 \times 10^{-4}$ is applied to the
encoder networks, and the latent loss weight is set to 2.0 with linear annealing
throughout training.

\paragraph{Experiment Settings}
In our experiments, we use the SMOKE5K dataset as the source domain dataset, and
the IJmond Camera dataset as the target domain dataset for pseudo-labeling. The
IJmond900 dataset is used for benchmarking the performance of our proposed
method. We evaluate our model on the IJmond900 dataset using the evaluation metrics
described in Section \ref{sec:Evaluation Metrics}. The complete experiment
settings are illustrated in Table \ref{tab:experiment_settings}, and the model symbols
and descriptions are shown in Table \ref{tab:model_dataset_symbols}.

\begin{table}[H]
    \centering
    \small
    \begin{tabular}{cccccc}
        \toprule ID               & Source Domain                    & Target Domain                   & Model                         & Pseudo-label Source                      & Constraint         \\
        \midrule$\mathcal{E}_{1}$ & $\mathcal{D}_{\mathrm{smoke5k}}$ & --                              & $\mathcal{M}_{\mathrm{full}}$ & --                                       & --                 \\
        $\mathcal{E}_{2}$         & $\mathcal{D}_{\mathrm{smoke5k}}$ & $\mathcal{D}_{\mathrm{ijmond}}$ & $\mathcal{M}_{\mathrm{weak}}$ & $\mathcal{M}_{\mathrm{smoke5k-full}}$    & Citizen Constraint \\
        $\mathcal{E}_{3}$         & $\mathcal{D}_{\mathrm{smoke5k}}$ & $\mathcal{D}_{\mathrm{ijmond}}$ & $\mathcal{M}_{\mathrm{DA}}$   & $\mathcal{M}_{\mathrm{smoke5k-full}}$    & No Constraint      \\
        $\mathcal{E}_{4}$         & $\mathcal{D}_{\mathrm{smoke5k}}$ & $\mathcal{D}_{\mathrm{ijmond}}$ & $\mathcal{M}_{\mathrm{DA}}$   & $\mathcal{M}_{\mathrm{smoke5k-full}}$    & Citizen Constraint \\
        $\mathcal{E}_{5}$         & $\mathcal{D}_{\mathrm{smoke5k}}$ & $\mathcal{D}_{\mathrm{ijmond}}$ & $\mathcal{M}_{\mathrm{DA}}$   & $\mathcal{M}_{\mathrm{smoke5k-full}}$    & Expert Constraint  \\
        $\mathcal{E}_{6}$         & $\mathcal{D}_{\mathrm{smoke5k}}$ & $\mathcal{D}_{\mathrm{ijmond}}$ & $\mathcal{M}_{\mathrm{DA}}$   & $\mathcal{M}_{\mathrm{ijmond900-train}}$ & Citizen Constraint \\
        \bottomrule
    \end{tabular}
    \caption{Core experimental settings. Note that the $\mathcal{E}_{6}$ has used
    pixel-wise annotations in the target domain as \textit{golden pseudo-label},
    indicating the model potential. The \emph{Constraint} column refers to citizen
    or expert feedback constraints on pseudo-label selection as described in Section~\ref{sec:Human-in-the-Loop
    Pseudo Label Refinement}.}
    \label{tab:experiment_settings}
\end{table}

\begin{table}[H]
    \centering
    \small
    \begin{adjustbox}
        {max width=\textwidth, center}
        \begin{tabular}{cl}
            \toprule Symbol                                                                       & Description                                                                 \\
            \midrule $\mathcal{M}_{\mathrm{smoke5k-full}}$                                        & TGB model pretrained on $\mathcal{D}_{\mathrm{smoke5k}}$ with full supervision\\
            $\mathcal{M}_{\mathrm{ijmond900-train}}$                                              & TGB model pretrained on $\mathcal{D}_{\mathrm{ijmond900-train}}$ with full supervision\\
            $\mathcal{M}_{\mathrm{full}}$~\cite{yan2023transmissionguidedbayesiangenerativemodel} & TGB model trained with full supervision                                     \\
            $\mathcal{M}_{\mathrm{weak}}$~\cite{MultiX-Amsterdam-ijmond-camera-ai}                & TGB model trained with weak-supervision and local contrastive loss          \\
            $\mathcal{M}_{\mathrm{DA}}$ (ours)                                                    & TGB model trained within our domain-adaptation framework                    \\
            $\mathcal{D}_{\mathrm{smoke5k}}$                                                      & SMOKE5K dataset (source domain, fully pixel-annotated)                      \\
            $\mathcal{D}_{\mathrm{ijmond}}$                                                       & IJmond Camera dataset (target domain, pseudo-labels), extracted from videos \\
            $\mathcal{D}_{\mathrm{ijmond900-train}}$                                              & 100 pixel-annotated images from IJmond900 (for golden standard training)    \\
            $\mathcal{D}_{\mathrm{ijmond900-test}}$                                               & 100 pixel-annotated images from IJmond900 (for testing only)                \\
            $\mathcal{E}_{i}$                                                                     & Experiment index                                                            \\
            \bottomrule
        \end{tabular}
    \end{adjustbox}
    \caption{Explanation of Model and Dataset Symbols.}
    \label{tab:model_dataset_symbols}
\end{table}


\chapter{Results} 


\ifpdf \graphicspath{{7/figures/PNG/}{7/figures/PDF/}{7/figures/}} \else
\graphicspath{{7/figures/EPS/}{7/figures/}} \fi

%
\section{Quantitative Comparison}
\label{sec:Quantitative Comparison} We hereby present the results of our
experiments in Table \ref{tab:experiment_results} according to the experiment
settings described in Table \ref{tab:experiment_settings}. The results demonstrate
the effectiveness of our proposed domain adaptation framework. Specifically, the
model with domain adaptation ($\mathcal{E}_{3}$) significantly outperforms the baseline
model ($\mathcal{E}_{2}$), with the $F_{1}$ score increasing from 0.083 to 0.387
and the $IoU_{smoke}$ from 0.043 to 0.240. This highlights the critical role of domain
adaptation in bridging the gap between the source and target domains.

Building on this, the introduction of citizen constraints further enhances
performance. Our main model ($\mathcal{E}_{4}$) achieves an $F_{1}$ score of 0.414
and an $IoU_{smoke}$ of 0.261, representing a five-fold and six-fold improvement
over the baseline ($\mathcal{E}_{2}$), respectively. This underscores the
powerful synergy between domain adaptation and citizen science. Furthermore, the
performance of $\mathcal{E}_{4}$ approaches that of the model with expert
constraints ($\mathcal{E}_{5}$, $F_{1}=0.448, IoU_{smoke}=0.288$), demonstrating
that citizen feedback can effectively approximate expert supervision while
enabling scalable annotation efforts by lowering the expertise barrier and
facilitating broader participation. While both citizen and expert labels are based
on video-level labels, involving citizen scientists allows for greater data
coverage and sustainability without substantially increasing the cost of
annotation.

We also observe that the model trained with large-scale source domain data and
citizen-constrained pseudo-labels ($\mathcal{E}_{4}$) achieves results ($F_{1}=0.
414, IoU_{smoke}=0.261$) comparable to the model trained with a small set of
fully annotated target domain data ($\mathcal{E}_{6}$,
$F_{1}=0.418, IoU_{smoke}=0.264$). This observation suggests that extensive
source domain annotation, when combined with our domain adaptation framework,
can match or even exceed the effectiveness of a limited amount of pixel-wise
annotated target domain data. This underscores the potential of domain adaptation
to significantly alleviate the high costs associated with manual annotation in the
industrial smoke detection scenario.

\begin{table}[htb]
    \centering
    \begin{tabular}{cccccc}
        \toprule Experiment                                                   & \multicolumn{5}{c}{ $\mathcal{D}_{\mathrm{ijmond900-test}}$} \\
        \midrule                                                              & $Recall$                                                    & $F_{1}$        & $IoU_{smoke}$  & $mIoU$         & $mMSE$         \\
        \cline{2-6} $\mathcal{E}_{1}$                                         & 0.234                                                       & 0.079          & 0.041          & 0.457          & 0.105          \\
        $\mathcal{E}_{2}$ (baseline~\cite{MultiX-Amsterdam-ijmond-camera-ai}) & \textbf{0.883}                                              & 0.083          & 0.043          & 0.287          & 0.406          \\
        $\mathcal{E}_{3}$                                                     & 0.557                                                       & 0.387          & 0.240          & 0.599          & 0.033          \\
        \textbf{$\mathcal{E}_{4}$}                                            & 0.527                                                       & 0.414          & 0.261          & 0.614          & 0.029          \\
        $\mathcal{E}_{5}$                                                     & 0.483                                                       & \textbf{0.448} & \textbf{0.288} & \textbf{0.630} & \textbf{0.024} \\
        $\mathcal{E}_{6}$ (golden standard)                                   & 0.460                                                       & 0.418          & 0.264          & 0.617          & 0.025          \\
        \bottomrule
    \end{tabular}
    \caption{Experiment results on $\mathcal{D}_{\mathrm{ijmond900-test}}$.}
    \label{tab:experiment_results}
\end{table}

To better evaluate the performance of smoke capturing, we compare the performance
of different experiment setups across varying levels of smoke opacity. As shown
in Table~\ref{tab:opacity_evaluation}, we observe that all models perform relatively
worse in low-opacity smoke than high-opacity smoke, which is as expected. This is
likely due to the inherent challenges of detecting low-opacity smoke, which
often has less distinct visual features and can be easily confused with
background noise or other environmental factors, underscoring the challenges of low-opacity
smoke segmentation in real-world industrial scenarios. The experiment with
domain adaptation and expert constraints ($\mathcal{E}_{5}$) achieves the best
performance in both high- and low-opacity scenarios, which is consistent with
the trends observed in Table~\ref{tab:experiment_results}.

\begin{table}
    \centering
    \begin{tabular}{ccccccc}
        \toprule Experiment                 & \multicolumn{3}{c}{High Opacity} & \multicolumn{3}{c}{Low Opacity} \\
        \midrule                            & $Recall$                         & $F_{1}$                        & $IoU_{high}$   & $Recall$       & $F_{1}$        & $IoU_{low}$    \\
        \cline{2-7} $\mathcal{E}_{1}$       & 0.400                            & 0.124                          & 0.075          & 0.187          & 0.065          & 0.038          \\
        $\mathcal{E}_{2}$ (baseline)        & \textbf{0.942}                   & 0.086                          & 0.048          & \textbf{0.867} & 0.099          & 0.058          \\
        $\mathcal{E}_{3}$                   & 0.829                            & 0.484                          & 0.356          & 0.480          & 0.415          & 0.297          \\
        \textbf{$\mathcal{E}_{4}$}          & 0.829                            & 0.527                          & 0.396          & 0.442          & 0.414          & 0.301          \\
        $\mathcal{E}_{5}$                   & 0.771                            & \textbf{0.568}                 & \textbf{0.439} & 0.402          & \textbf{0.460} & \textbf{0.343} \\
        $\mathcal{E}_{6}$ (golden standard) & 0.748                            & 0.557                          & 0.428          & 0.379          & 0.437          & 0.320          \\
        \bottomrule
    \end{tabular}
    \caption{Performance on $\mathcal{D}_{\mathrm{ijmond900-test}}$ for the
    different levels of opacity. }
    \label{tab:opacity_evaluation}
\end{table}

\section{Qualitative Comparison on IJmond900 Dataset}
\label{sec:Qualitative Comparison on IJmond900 Dataset} 
\begin{figure}[htb]
    \centering
    \begin{minipage}[t]{0.24\textwidth}
        \centering
        \includegraphics[width=\linewidth]{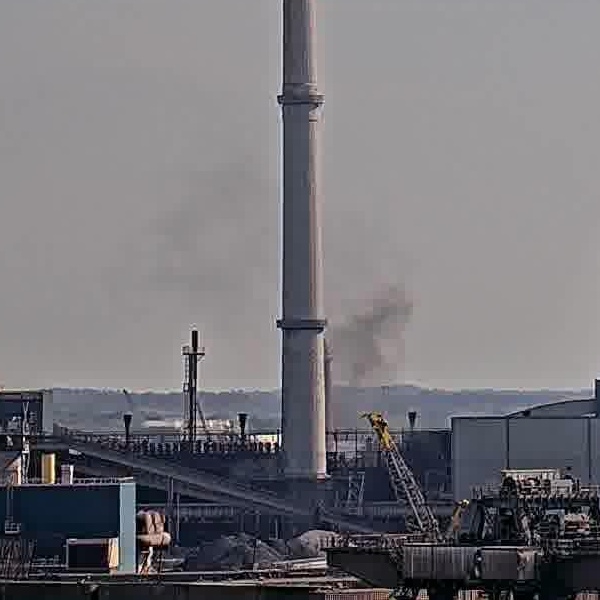}
        \par
        \vspace{0.5ex}
        \small (a)
    \end{minipage}
    \hfill
    \begin{minipage}[t]{0.24\textwidth}
        \centering
        \includegraphics[width=\linewidth]{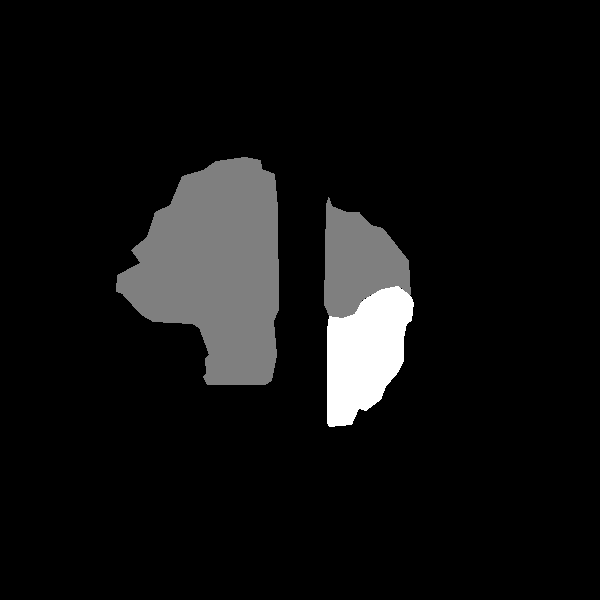}
        \par
        \vspace{0.5ex}
        \small (b)
    \end{minipage}
    \hfill
    \begin{minipage}[t]{0.24\textwidth}
        \centering
        \includegraphics[width=\linewidth]{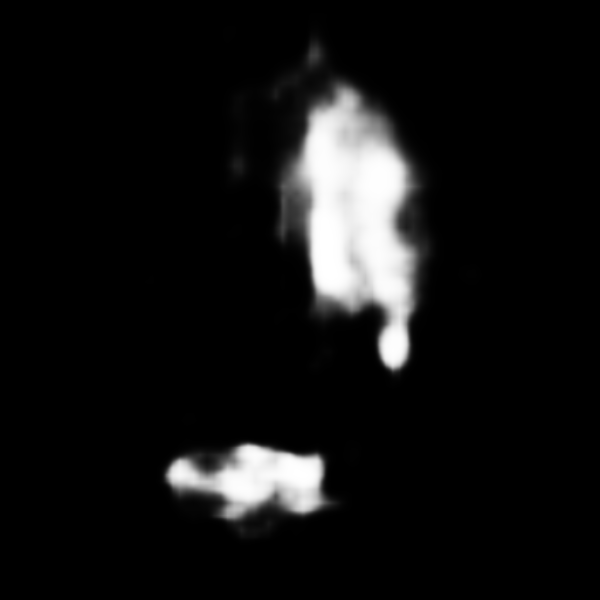}
        \par
        \vspace{0.5ex}
        \small (c)
    \end{minipage}
    \hfill
    \begin{minipage}[t]{0.24\textwidth}
        \centering
        \includegraphics[width=\linewidth]{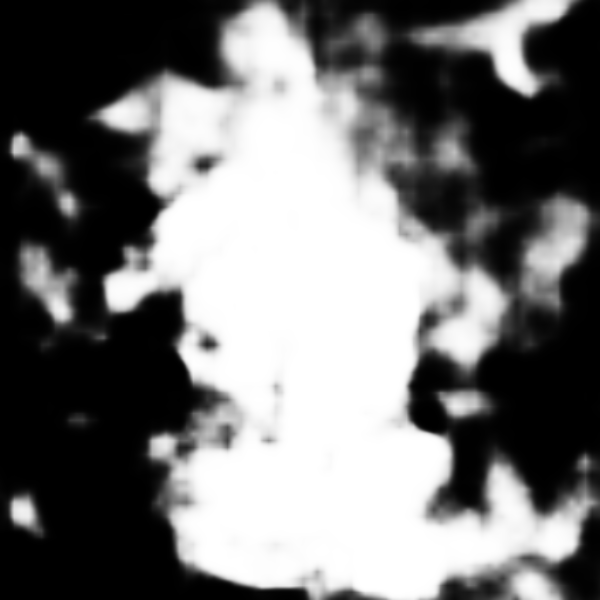}
        \par
        \vspace{0.5ex}
        \small (d)
    \end{minipage}

    \vspace{1.2ex}

    \begin{minipage}[t]{0.24\textwidth}
        \centering
        \includegraphics[width=\linewidth]{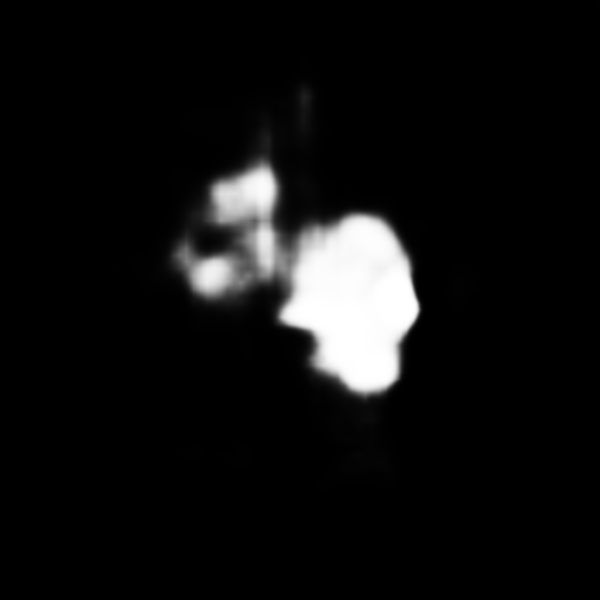}
        \par
        \vspace{0.5ex}
        \small (e)
    \end{minipage}
    \hfill
    \begin{minipage}[t]{0.24\textwidth}
        \centering
        \includegraphics[width=\linewidth]{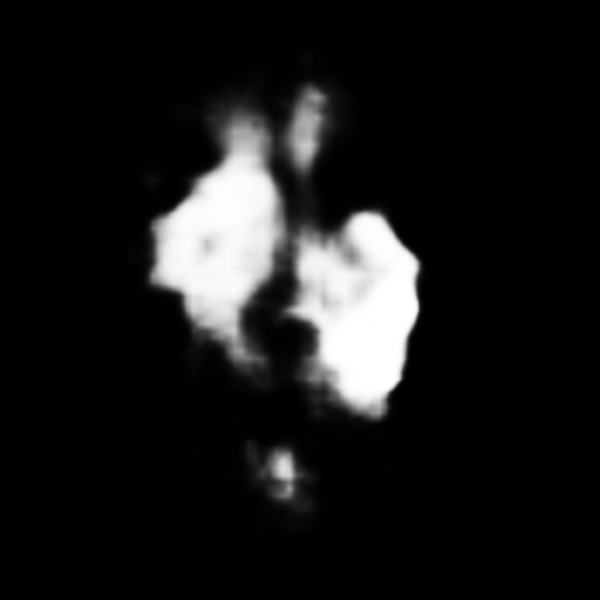}
        \par
        \vspace{0.5ex}
        \small (f)
    \end{minipage}
    \hfill
    \begin{minipage}[t]{0.24\textwidth}
        \centering
        \includegraphics[width=\linewidth]{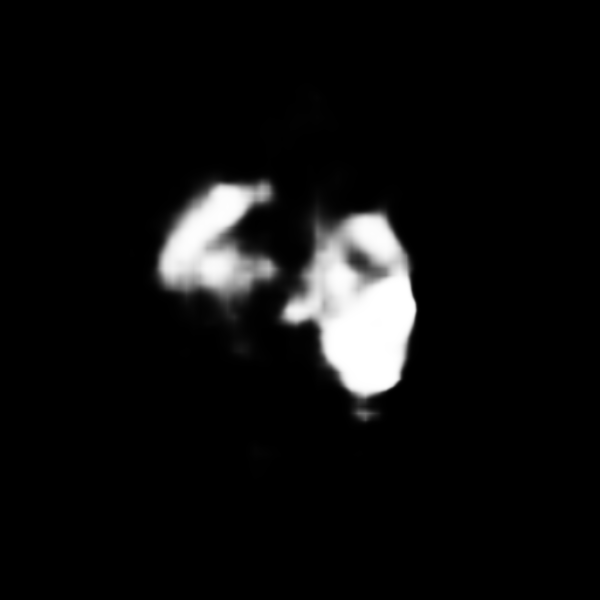}
        \par
        \vspace{0.5ex}
        \small (g)
    \end{minipage}
    \hfill
    \begin{minipage}[t]{0.24\textwidth}
        \centering
        \includegraphics[width=\linewidth]{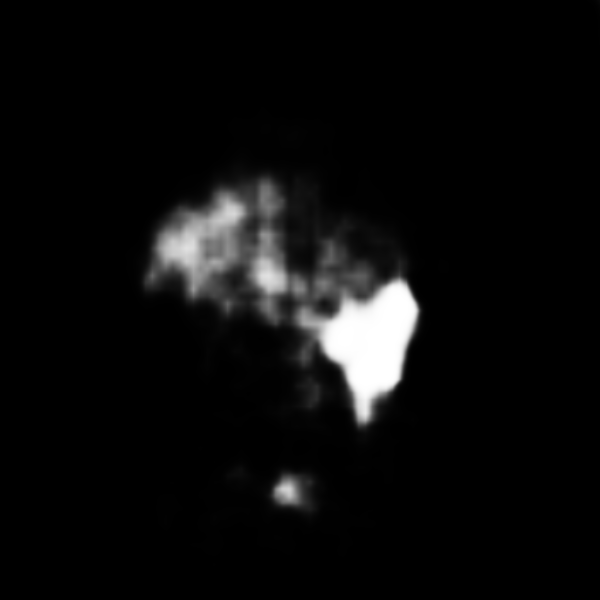}
        \par
        \vspace{0.5ex}
        \small (h)
    \end{minipage}

    \caption{The qualitative results of different models on
    $\mathcal{D}_{\mathrm{ijmond900-test}}$. From left to right, top to bottom: (a)
    Input image, (b) Ground truth, (c) Model $\mathcal{E}_{1}$, (d) Model
    $\mathcal{E}_{2}$, (e) Model $\mathcal{E}_{3}$, (f) Model $\mathcal{E}_{4}$,
    (g) Model $\mathcal{E}_{5}$, (h) Model $\mathcal{E}_{6}$. The models correspond
    to the experiments listed in Table~\ref{tab:experiment_settings}.}
    \vspace{1ex}
    \label{fig:qualitative-results}
\end{figure}

We also visualize the predictions
of our different models in Fig.~\ref{fig:qualitative-results}.

We find that our method produces more accurate predictions than the baseline model,
particularly in high-opacity background regions. The model trained with citizen constraints
achieves a favorable balance between precision and recall, effectively
identifying smoke regions while minimizing misalignments. Incorporating expert constraints
yields a further increase in $F_{1}$ and $IoU_{smoke}$, suggesting that expert
supervision can help refine segmentation boundaries and reduce false positives. However,
we also observe that the presence of steam often disrupts smoke detection, as the
model tends to misclassify steam-like objects as smoke, which is especially
pronounced in videos containing extensive steam presence and positive labels. This
confusion may arise because steam generally exhibits stronger temporal and spatial
consistency compared to smoke emissions.

Overall, our domain adaptation framework, especially when combined with human-in-the-loop
pseudo-label refinement, significantly improves the segmentation accuracy and
robustness in complex industrial scenarios. This validates the practical value of
leveraging large-scale weak labels from non-experts for real-world toxic smoke emission
tasks.

\section{Ablation Study}
\label{sec:Ablation Study} To analyze each component of our framework, we
conduct an ablation study to investigate the effects of Attention Pooling, LDConv,
and human feedback pseudo-label constraints on model performance. These components
are chosen because prior studies show attention pooling improves class-specific feature
aggregation, LDConv handles spatial deformations, and human feedback reduces pseudo-label
noise \cite{paul2020domainadaptivesemanticsegmentation, Zhang_2024, Kosmala2016}.
Results are illustrated in Table~\ref{tab:ablation_study}.

\begin{table}[htb]
    \centering
    \begin{tabular}{cccccll}
        \toprule $Base$     & $AttPool$  & $LDConv$   & Constraint & $Recall$       & $F_{1}$        & $IoU_{smoke}$  \\
        \midrule \checkmark & \checkmark & \checkmark & No         & 0.557          & 0.387          & 0.240          \\
        \checkmark          &            &            & Citizen    & 0.402          & 0.395          & 0.246          \\
        \checkmark          & \checkmark &            & Citizen    & 0.411          & 0.392          & 0.243          \\
        \checkmark          &            & \checkmark & Citizen    & \textbf{0.563} & 0.401          & 0.251          \\
        \checkmark          & \checkmark & \checkmark & Citizen    & 0.527          & 0.414          & 0.261          \\
        \checkmark          & \checkmark & \checkmark & Expert     & 0.483          & \textbf{0.448} & \textbf{0.288} \\
        \bottomrule
    \end{tabular}
    \caption{Ablation study for the contributions of LDConv, attention pooling,
    and human feedback pseudo-label constraints.}
    \label{tab:ablation_study}
\end{table}

\subsection{Impact of Attention Pooling}
Our ablation study reveals that the impact of attention pooling is nuanced and highly
dependent on its interaction with other components. When used in isolation
alongside the citizen-constrained baseline, attention pooling provides a marginal
increase in recall (from 0.402 to 0.411) but at the cost of slightly lower $F_{1}$-score
and $IoU_{smoke}$. This suggests that, on its own, it struggles to meaningfully
improve overall segmentation quality but make the model more perculiar in its
predictions.

However, its contribution becomes more apparent when combined with LDConv.
Adding attention pooling to the LDConv-enhanced model causes recall to drop from
0.563 to 0.527, but it simultaneously boosts the $F_{1}$-score from 0.401 to 0.414
and $IoU_{smoke}$ from 0.251 to 0.261. This indicates that attention pooling
helps refine the features captured by LDConv, making the model more precise by focusing
on high-confidence predictions, albeit at the expense of detecting some less
obvious smoke regions. It trades higher recall for better overall segmentation accuracy.

\subsection{Impact of LDConv}
The inclusion of LDConv provides the most significant performance boost among the
architectural components, particularly in terms of recall. As shown in Table~\ref{tab:ablation_study},
adding only LDConv to the citizen-constrained baseline improves the recall from
0.402 to \textbf{0.563} --- the highest recall value achieved under citizen constraints.
This demonstrates LDConv's powerful ability to capture a wide array of smoke-like
features and handle the high variability in smoke's appearance.

This substantial improvement can be attributed to LDConv's capacity for modeling
fine-grained local context and adapting to spatial deformations. By dynamically aggregating
local features, LDConv equips the model to identify subtle or diffuse smoke
patterns that other configurations might miss, thus maximizing detection sensitivity.
While it also provides a modest lift to $F_{1}$-score and $IoU_{smoke}$, its primary
contribution is clearly in enhancing the model's ability to find potential smoke
regions. The combination of LDConv with attention pooling further refines this
ability, allowing the model to focus on high-confidence detections while still maintaining
a strong recall, providing complementary strengths to the overall model
performance.

\subsection{Impact of Human Feedback Constraints}
Introducing human feedback constraints, whether from citizens or professional researchers,
consistently enhances model performance by improving the quality of pseudo-labels
used for training. As shown in Table~\ref{tab:ablation_study}, when both LDConv
and attention pooling are present, applying citizen constraints increases $F_{1}$
from 0.387 (no constraint) to 0.414, while expert constraints further raise
$F_{1}$ to 0.448 and $IoU_{smoke}$ to 0.288, representing the best performance among
all test configurations.

This performance boost can be attributed to the fact that human feedback, even
when provided by non-experts, helps to filter out grossly incorrect pseudo-labels
and guides the model towards more meaningful smoke annotations. Citizen
constraints supply a broader but slightly noisier supervision signal, improving recall
and model robustness, whereas expert constraints yield more precise and reliable
feedback, resulting in fewer false positives. Importantly, the consistent improvement
from citizen to expert feedback underscores the value of human-in-the-loop approaches,
particularly in scenarios where fully annotated target domain data are scarce or
expensive to obtain.

\subsection{Component Synergy and Trade-offs}
\label{sec:Component Synergy and Trade-offs} An interesting insight from our
ablation study is the trade-off between recall and precision, which highlights the
nuanced interplay between different model components. For instance, the configuration
with only LDConv and citizen constraints achieves the highest recall (0.563),
suggesting it is highly effective at identifying potential smoke regions. However,
its $F_{1}$-score (0.401) is surpassed by the full model with citizen
constraints (0.414), which has a lower recall (0.527).

This reveals a classic trade-off: LDConv excels at capturing a wide range of
smoke-like features, maximizing detection sensitivity (high recall), but
potentially at the cost of including less certain or noisy predictions. On the other
hand, adding attention pooling makes the model more conservative and precise. It
learns to refine the feature map and focus on high-confidence regions, thereby
improving the overall segmentation quality ($F_{1}$ and $IoU_{smoke}$) but
slightly reducing its ability to detect every possible smoke instance. This finding
underscores that the optimal model configuration may depend on the specific application's
priority --- whether maximizing detection rates or ensuring the highest possible
segmentation accuracy.


\chapter{Discussions} 


\ifpdf \graphicspath{{7/figures/PNG/}{7/figures/PDF/}{7/figures/}} \else
\graphicspath{{7/figures/EPS/}{7/figures/}} \fi

%

In this chapter, we discuss some key findings from our experiments and their
implications for future research in industrial smoke detection in the context of
WDA and citizen science approaches.

\section{Domain Gap and Pseudo-label Generation Strategies}
\label{sec:Domain Gap and Pseudo-label Generation Strategies}

\begin{figure}[htb]
    \centering
    \begin{minipage}[t]{0.9\textwidth}
        \centering
        \includegraphics[height=3.3cm]{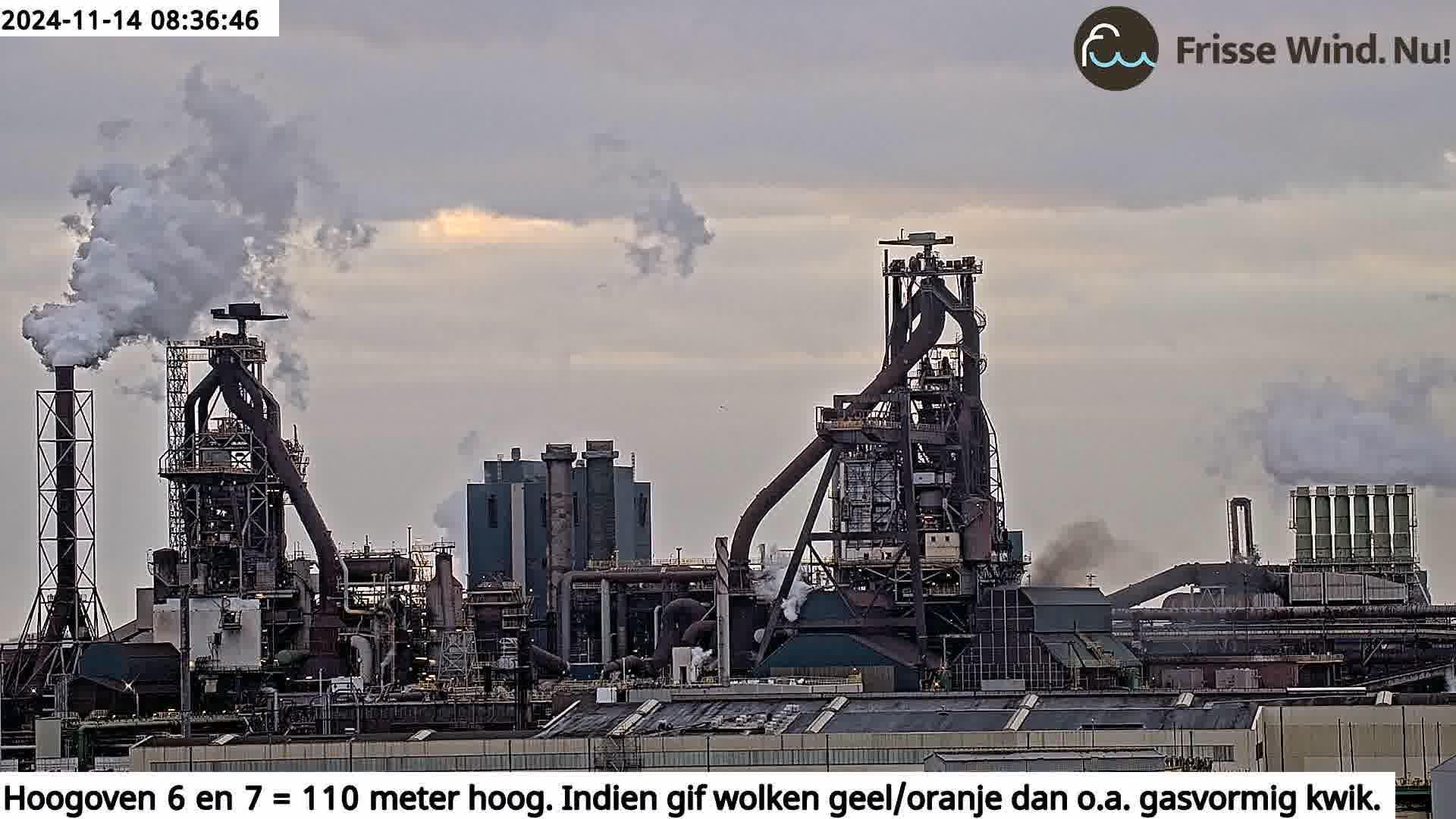}
        \includegraphics[height=3.3cm]{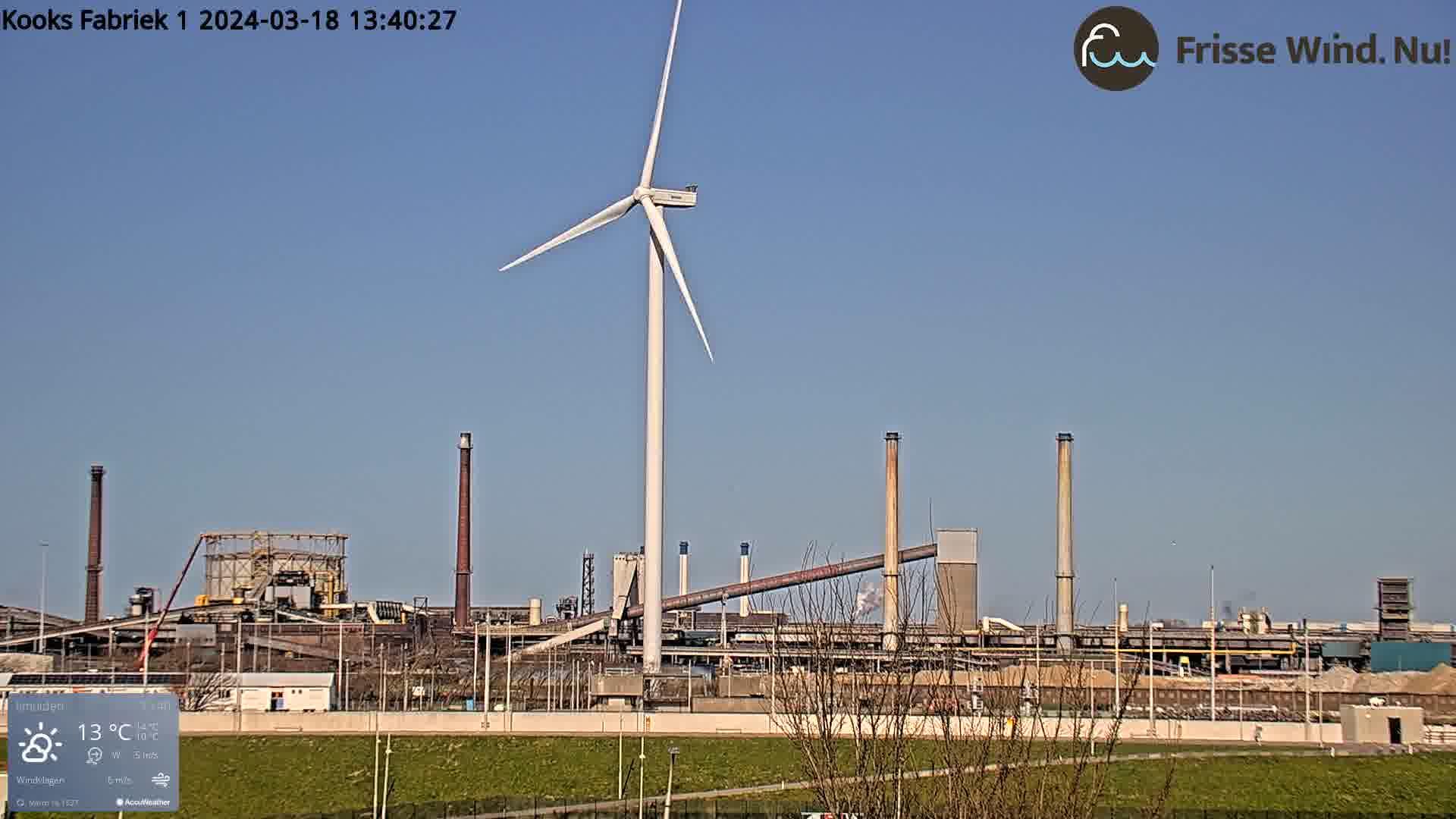}
        \\[-1ex]
        {(a) Images from IJmond900 dataset}
    \end{minipage}

    \vspace{1.2ex}
    \begin{minipage}[t]{0.46\textwidth}
        \centering
        \includegraphics[height=3.3cm]{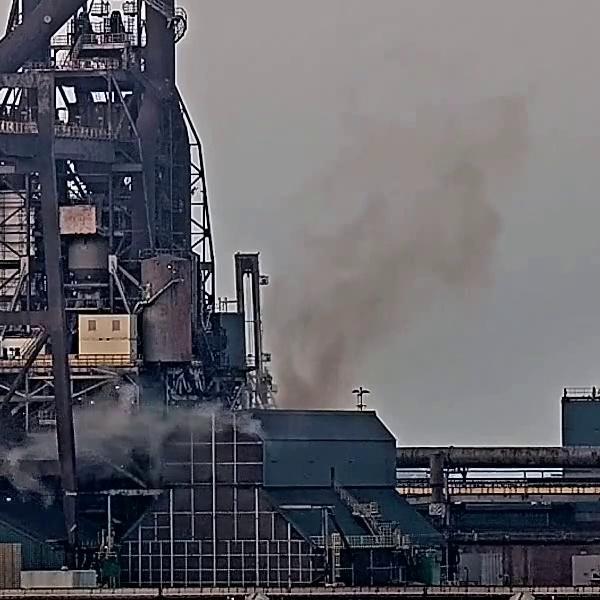}
        \includegraphics[height=3.3cm]{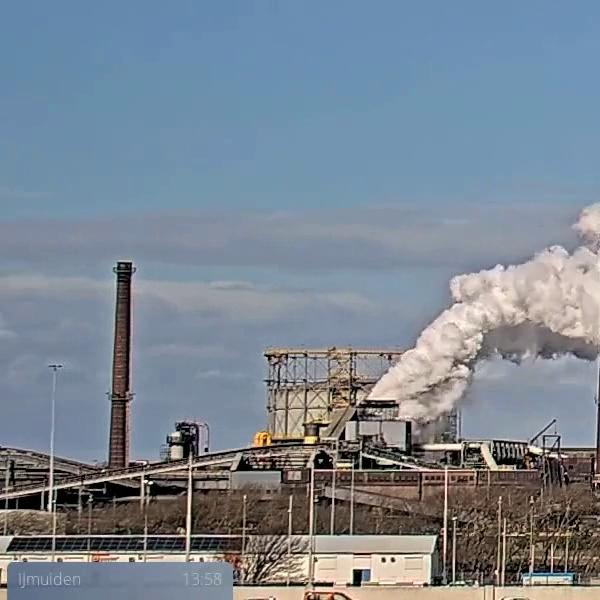}
        \\[-1ex]
        {(b) Images from IJmond Camera dataset}
    \end{minipage}%
    \hfill
    \begin{minipage}[t]{0.54\textwidth}
        \centering
        \includegraphics[height=3.3cm]{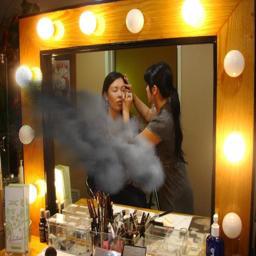}
        \includegraphics[height=3.3cm]{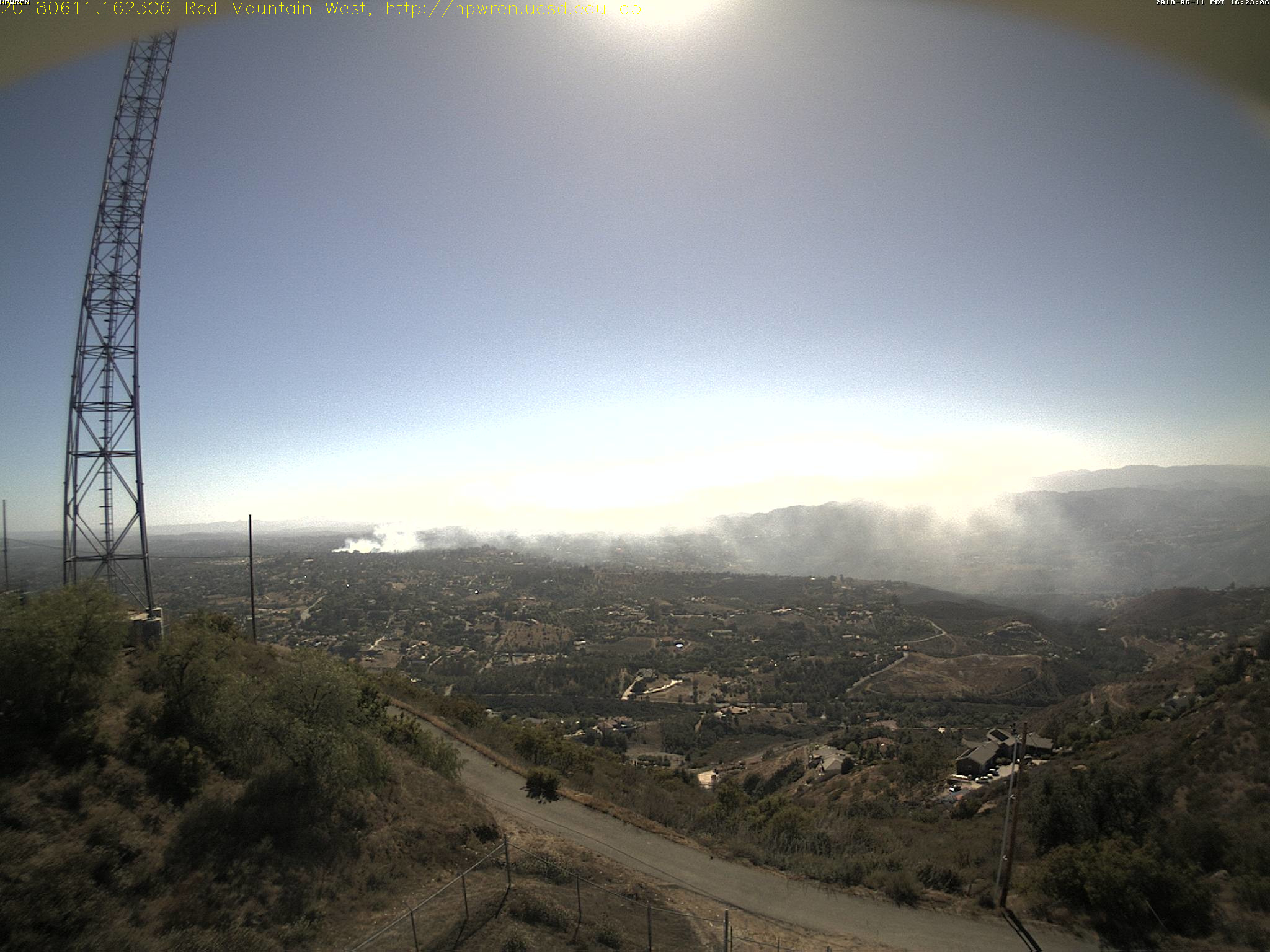}
        \\[-1ex]
        {(c) Images from SMOKE5K dataset}
    \end{minipage}

    \caption{Domain gap between source and target domains. (a) Images from the \textbf{IJmond900}
    dataset, captured from three different cameras at a later period. (b) Images
    from the \textbf{IJmond Camera} dataset, which are earlier, cropped versions
    from the same panoramic views. (c) Images from the \textbf{SMOKE5K} dataset.
    Left: synthetic image; right: real-world wildfire footage.}
    \label{fig:datasets_description}
\end{figure}

While all datasets we used in this thesis aim at smoke detection, they differ
markedly in appearance and capture settings, as illustrated in Figure~\ref{fig:datasets_description}.
The \textbf{IJmond900} dataset comprises frames from three distinct cameras in the
IJmond area, all acquired at later dates. By contrast, the \textbf{IJmond Camera}
dataset holds earlier frames, which are cropped from the same panoramic views of
IJmond900. This temporal offset and multi-camera acquisition introduce a clear and
inevitable domain gap. Environmental factors, such as season, weather, lighting,
and time of day, further increase visual heterogeneity, resulting in a highly
heterogeneous distribution of target domain data.

This variability surpasses the relatively controlled domain shifts found in well-studied
benchmark datasets such as SYNTHIA~\cite{Ros2016}, GTA5~\cite{richter2016playingdatagroundtruth},
and Cityscapes~\cite{cordts2016cityscapesdatasetsemanticurban}, where scenes and
classes are tightly controlled for experimental comparability. In those settings,
pseudo-labeling usually generalizes effectively. However, in real industrial scenarios,
the gap between synthetic training data and real factory footage is both large
and understudied. Transparent, amorphous plumes and steam-like artifacts make
accurate pseudo-labeling extremely challenging even for human experts. These challenges
motivate our use of class-aware WDA augmented by citizen-science feedback, which
narrows the gap and markedly enhances pseudo-label quality for industrial smoke
segmentation.

\section{The Power of Source Data: A Case for Domain Adaptation}
\label{sec:The Power of Source Data: A Case for Domain Adaptation}

An intriguing and pivotal finding from our experiments is that the model trained
with large-scale source data and citizen-refined pseudo-labels ($\mathcal{E}_{4}$)
achieved performance nearly on par with the model trained on a small, fully-annotated
target domain dataset ($\mathcal{E}_{6}$). Specifically, $\mathcal{E}_{4}$ achieved
an $F_{1}$-score of 0.414 and an $IoU_{smoke}$ of 0.261, which is remarkably
close to the performance of $\mathcal{E}_{6}$ ($F_{1}=0.418, IoU_{smoke}=0.264$).
This result is a powerful testament to the value of domain adaptation.

This observation underscores a critical conclusion: vast and diverse source
domain data, when leveraged through effective domain adaptation, can provide a
feature representation nearly as robust and generalizable as that from small, perfectly
labeled target domain data. The limited size of the target training set (100 images),
while high-quality, may not be sufficient to capture the full spectrum of visual
variations present in the complex industrial environment of IJmond. Factors such
as changing weather, lighting conditions, and diverse smoke characteristics demand
a far larger dataset to model effectively.

Consequently, this finding strongly validates the core premise of our research. It
highlights the immense potential and necessity of domain adaptation for industrial
smoke segmentation. Rather than investing prohibitive effort and cost into
manually annotating thousands of target images, domain adaptation offers a pragmatic
and scalable alternative. By effectively bridging the domain gap, we can harness
the rich knowledge embedded in large, readily available source datasets (such as
synthetic data) and transfer it to a target domain where labeled data is scarce.
This not only makes the problem tractable but also paves the way for developing highly
effective monitoring systems in real-world industrial settings. Thus, the comparable
performance of the domain-adapted model reinforces the argument that advancing domain
adaptation techniques is a crucial and highly valuable research direction for this
application.

\section{Strengths, Limitations, and Value of Human Feedback}
\label{sec:Strengths, Limitations, and Value of Human Feedback}

We assess model performance using Recall, $F_{1}$-score, IoU of smoke, mIoU, and
mMSE. For industrial smoke monitoring, high Recall is crucial to avoid missing
hazardous events, but often leads to reduced Precision and more false alarms.

Our results show that incorporating human feedback into class-aware domain adaptation
improves Recall, Precision and IoU, while reducing mMSE compared to unsupervised adaptation.
Citizen-provided weak labels are effective in reducing the domain gap and enhancing
segmentation quality. Although expert constraints further improve Precision and
mIoU, they bring limited Recall gains, indicating that precise boundaries alone
cannot fully resolve the domain shift for smoke-like vague objects. Golden pseudo-labels
confirm that limited expert annotation can boost segmentation accuracy, but such
approaches are impractical for large-scale deployment due to high cost.

Crucially, our study demonstrates the practical value of scalable, citizen-driven
weak supervision: it enables robust, adaptive smoke segmentation with far less
annotation effort, making continuous large-scale monitoring feasible. This lays
the groundwork for combining community feedback with advanced weak- and active learning
in the future.

However, given the considerable domain gap between synthetic source data and
real-world target data, and the weaker annotation quality in the target domain,
overall segmentation performance remains moderate. There is also a residual gap between
the weakly labeled target domain and the fully annotated test set, which
complicates evaluation. Still, our results align with previous studies on weakly
supervised domain adaptation~\cite{Das2023, Wang_2019}, which also report limited
performance gains when adapting from synthetic to real-world data, indicating
that domain adaptation remains a challenging task, especially in industrial contexts
where visual variability is high.

Notably, our research is the first to combine domain adaptation, citizen science,
and apply them in this specific industrial toxic emission detection task. Unlike standard datasets
featuring salient, well-defined objects (e.g., vehicles, pedestrians),
industrial smoke is small, transparent, and variable in shape, often blending with
background elements like steam. These unique challenges make pixel-level
segmentation particularly demanding, even for fully supervised methods. Thus, while
our absolute metrics may not match those of conventional benchmarks, the
progress in tackling real-world industrial challenges and the consistent
relative improvements are of greater significance. The robust gains over baseline
validate the value of our approach.


\chapter{Conclusion} 


\ifpdf \graphicspath{{7/figures/PNG/}{7/figures/PDF/}{7/figures/}} \else
\graphicspath{{7/figures/EPS/}{7/figures/}} \fi

%

In this thesis, we proposed CEDANet, a novel human-in-the-loop domain adaptation
framework for industrial smoke segmentation. Its core innovation lies in the
synergistic integration of class-aware domain adaptation with weakly-supervised
constraints derived from citizen science. Through comprehensive experiments on the
SMOKE5K and custom real-world datasets, we demonstrated that CEDANet can
effectively mitigate the significant domain shift between synthetic and target
industrial data, achieving superior segmentation accuracy with minimal reliance on
expert annotations. Notably, our results showed that a model adapted with large-scale
source data and citizen-informed pseudo-labels could match or even exceed the performance
of one trained on a small-scale, fully-annotated target set. This underscores the
scalability and cost-efficiency of our proposed pipeline. Our findings confirm
that leveraging community feedback within a robust domain adaptation framework
is a powerful and practical strategy for complex environmental monitoring tasks,
paving the way for future investigations into active learning and other applications
facing data scarcity.



\chapter{Appendix} 
\setcounter{equation}{0}
\renewcommand{\theequation}{A.\arabic{equation}} 
\setcounter{figure}{0}
\setcounter{section}{0}
\renewcommand{\thesection}{\Alph{section}} 


\ifpdf \graphicspath{{X/figures/PNG/}{X/figures/PDF/}{X/figures/}} \else
\graphicspath{{X/figures/EPS/}{X/figures/}} \fi

\section{Bayesian Generative Model}
\subsection{Bayesian Statistics}
The very basic idea of Bayesian statistics is to update our beliefs about the
world based on the observation of new data. In a Bayesian framework, we start
with a prior distribution that represents our beliefs before observing any new
data. As we collect data, we update this prior to form a posterior distribution
using Bayes' theorem:
\begin{equation}
    P(H|D)\,=\,\frac{P(D|H)P(H)}{P(D)},
\end{equation}
where $P(H|D)$ is the posterior probability of hypothesis $H$ given data $D$, and
$P(D|H)$ is the likelihood of observing data $D$ given hypothesis $H$. The term
$P(H)$ is the prior probability of hypothesis $H$, and $P(D)$ is the marginal likelihood
of data $D$.
\subsection{Generative Models}
Bayesian generative models are a class of probabilistic models that aim to
capture the underlying structure of the data by modeling how the data is
generated. These models assume that the data is generated from a set of latent variables,
which are not directly observed but can be inferred from the data. In smoke segmentation
tasks, translucent edges, image noise, labeling ambiguities, etc., can lead to different
segmentation results for the same pixel in the same image, even if the same
model is fed multiple times. Such "natural jitter" is common in real-world scenarios
due to the data-level uncertainty. \\
We introduce the latent variables $z$ to model the aleatoric uncertainty in the
data generation process. The traditional deterministic network is constrained to
the single probabilistic output $\hat{s}= f_{\theta}(x)$, which cannot capture
the uncertainty in the data. Hence, latent variables $z$ are introduced to
capture "the jitter that the network cannot do anything about". With the latent variables,
the generative model can output different $\hat{s}$ for the same input $x$ with
different $z$ by different sampling strategies, generating a distribution of segmentation
results for the intrinsic uncertainty of the data. \\
A typical predictive distribution of Bayesian generative model can be expressed
as:
\begin{equation}
    P(s|x)\,=\,\int P(s|x,z)P(z|x)dz,
\end{equation}
where $P(z)$ is the prior distribution of the latent variables, and $P(s|x,z)$
indicates the likelihood of the segmentation result $s$ given the input image
$x$ and the latent variables $z$, and $P(z|x)$ is the posterior distribution of the
latent variables given the input image $x$, which can be expressed as:
\begin{equation}
    P(z|x)\,=\,\frac{P(x|z)P(z)}{P(x)}\,=\,\frac{P(x|z)P(z)}{\int P(x|z)P(z)dz},
\end{equation}
where $P(x|z)$ is the likelihood of the input image $x$ given the latent variables
$z$, and $P(z)$ is the prior distribution of the latent variables. The integral in
the denominator is the marginal likelihood of the input image $x$, which is
often intractable to compute directly due to the high dimensionality of $z$. Therefore,
we need to use approximate inference methods to estimate the posterior
distribution of the latent variables $z$ given the input image $x$.
\subsection{Variational Inference}
Variational inference is a powerful technique for approximating the posterior distribution
of latent variables in Bayesian generative models. The key idea is to introduce
a family of variational distributions $P(z|x)$ that approximates the true
posterior distribution $P(z|x)$. The goal is to find the variational distribution
that minimizes the Kullback-Leibler (KL) divergence between the true posterior and
the variational distribution:
\begin{equation}
    D_{KL}(Q(z|x)||P(z|x))\,=\,\int Q(z|x) \log \frac{Q(z|x)}{P(z|x)}dz,
\end{equation}
This can be rewritten using Bayes' theorem:
\begin{equation}
    D_{KL}(q(z|x)||p(z|x))\,=\,\int Q(z|x) \log \frac{Q(z|x)P(x)}{P(x|z)P(z)}dz,
\end{equation}
This leads to the evidence lower bound (ELBO):
\begin{equation}
    \mathcal{L}(Q)\,=\,\int Q(z|x) \log P(x|z)dz\,-\,D_{KL}(Q(z|x)||P(z)),
\end{equation}
The ELBO can be interpreted as a trade-off between the likelihood of the data given
the latent variables $z$ and the complexity of the variational distribution. By
maximizing the ELBO, we can find the optimal variational distribution $Q_{\theta^*}
(z|x)$ that approximates the true posterior distribution $P(z|x)$. This allows
us to perform approximate inference in Bayesian generative models, enabling us to
capture the uncertainty in the data generation process and generate a distribution
of segmentation results for the intrinsic uncertainty of the data.

\subsection{VAE and CVAE}
Variational Autoencoders (VAEs) and Conditional Variational Autoencoders (CVAEs)
are two popular types of Bayesian generative models that use variational inference
to approximate the posterior distribution of latent variables. VAEs are a type
of generative model that learns a latent representation of the data by encoding
the data into a lower-dimensional space and then decoding it back to the
original data space. The key idea is to learn a probabilistic mapping from the data
space to the latent space, allowing us to sample from the latent space and
generate new data points. The VAE consists of an encoder network that maps the input
data to a latent representation, and a decoder network that maps the latent
representation back to the data space. The encoder network outputs the parameters
of a probabilistic distribution (usually Gaussian) in the latent space, and the
decoder network outputs the parameters of a probabilistic distribution in the
data space. CVAEs extend VAEs by conditioning the generative process on additional
information, such as class labels or other auxiliary data. This allows the model
to generate data points that are conditioned on the additional information,
enabling the model to capture more complex relationships between the data and
the latent variables. The CVAE consists of a conditional encoder network that maps
the input data and the additional information to a latent representation, and a
conditional decoder network that maps the latent representation and the
additional information back to the data space. The conditional encoder network outputs
the parameters of a probabilistic distribution in the latent space, and the
conditional decoder network outputs the parameters of a probabilistic
distribution in the data space. The CVAEs can be trained using the same variational
inference framework as VAEs, by maximizing the evidence lower bound (ELBO) with
respect to the parameters of the encoder and decoder networks. This allows the model
to learn a probabilistic mapping from the data space to the latent space, and
from the latent space back to the data space, enabling the model to capture the uncertainty
in the data generation process and generate a distribution of segmentation results
for the intrinsic uncertainty of the data. The CVAE can be expressed as:
\begin{equation}
    P(s|x,y)\,=\,\int P(s|x,y,z)P(z|x,y)dz,
\end{equation}
where $y$ is the additional information that conditions the generative process,
$P (z|x,y)$ is the posterior distribution of the latent variables given the
input image $x$ and the additional information $y$, and $P(s|x,y,z)$ is the likelihood
of the segmentation result $s" given the input image$x", the additional
information $y", and the latent variables$z". The posterior distribution of the latent
variables can be expressed as:
\begin{equation}
    P(z|x,y)\,=\,\frac{P(x,y|z)P(z)}{P(x,y)}\,=\,\frac{P(x,y|z)P(z)}{\int P(x,y|z)P(z)dz}
    ,
\end{equation}
where $P(x,y|z)$ is the likelihood of the input image $x$ and the additional
information $y$ given the latent variables $z$, and $P(z)$ is the prior distribution
of the latent variables. The integral in the denominator is the marginal
likelihood of the input image $x$ and the additional information $y$, which is often
intractable to compute directly due to the high dimensionality of $z$. Therefore,
we need to use approximate inference methods to estimate the posterior distribution
of the latent variables $z$ given the input image $x$ and the additional information
$y$.

\subsection{Further on ELBO}
The ELBO is defined as the expectation of the log likelihood of the data under the
variational distribution minus the KL divergence between the variational distribution
and the prior distribution. Mathematically, it can be expressed as:
\begin{equation}
    \mathcal{L}(\theta, \omega)\,=\,\mathbb{E}_{Q_\omega(z|x)}[\log P_{\theta}(x
    | z) ]\,-\,D_{KL}(Q_{\omega}(z|x) || P(z)),
\end{equation}
where $Q_{\omega}(z|x)$ is the variational distribution, $P_{\theta}(x|z)$ is the
likelihood of the data given the latent variables, and $P(z)$ is the prior
distribution over the latent variables $z$. Moreover, the log distribution of
the data $x$ can be described as:
\begin{equation}
    \log{P_\theta(x)}\,=\,\mathcal{L}(\theta, \omega)\,+\,D_{KL}(Q_{\omega}(z|x)
    || P_{\theta}( z|x)),
\end{equation}
The second term of A.2, $D_{KL}(Q_{\omega}(z|x) || P_{\theta}(z|x))$, is non-negative,
and only when $Q_{\omega}(z|x) = P_{\theta}(z|x)$, the bound becomes tight, and maximizing
ELBO is equivalent to maximizing the true marginal likelihood. Therefore, $\mathcal{L}$is
indeed a lower bound on the marginal log-likel ihood of the data$x$. Furthermore,
maximizing$\mathcal{L}$ simultaneously:
\begin{itemize}
    \item rases the likelihood of the data $P_{\theta}$ under the variational distribution,

    \item pushes the variational distribution $Q_{\omega}$ closer to the true posterior
        distribution $P_{\theta}(z|x)$.
\end{itemize}
We treat the first term of A.1 as the reconstruction reward, and the second term
as the regularization loss. The first term rewards latent variables that
faithfully reconstruct the input, while the second one keeps them from drifting too
far away from the prior.\\
In practice we can use the reparameterization trick:
\begin{equation}
    z\,=\,\mu_{\omega}(x)\,+\,\sigma_{\omega}(x) \odot \epsilon,\quad \epsilon \sim
    \mathcal{N}(0, I),
\end{equation}
turning the stochastic expectation into a differentiable path and enabling end-to-end
training with stochastic gradient descent (SGD) techniques.

\section{Evaluation Metrics}
\label{Appendix:2}
\setcounter{equation}{0}
\renewcommand{\theequation}{B.\arabic{equation}} 

To provide a comprehensive and transparent evaluation of our smoke segmentation
models, all metrics are computed on a \textbf{pixel-wise basis}. The final reported
scores represent the average values across all images in the test set. Our
evaluation framework is designed to assess both overall binary segmentation
accuracy and the model's specific performance on smoke regions of varying
opacity.

Let $P$ be the set of pixels predicted as smoke by the model, and $G$ be the set
of all ground-truth smoke pixels. The ground truth is further divided into high-opacity
($G_{high}$) and low-opacity ($G_{low}$) regions, such that $G = G_{high}\cup G_{low}$.
The set of ground-truth background pixels is denoted by $B$.

\subsection{Overall Performance Metrics (Binary Smoke vs. Background)}
These metrics evaluate the model's ability to distinguish smoke from background
pixels. The standard pixel-wise definitions for True Positives (TP), False
Positives (FP), and False Negatives (FN) are used:
\begin{itemize}
    \item \textbf{True Positives (TP)}: $|P \cap G|$, smoke pixels correctly
        classified as smoke.

    \item \textbf{False Positives (FP)}: $|P \cap B|$, background pixels
        incorrectly classified as smoke.

    \item \textbf{False Negatives (FN)}: $|G| - |P \cap G|$, smoke pixels
        incorrectly classified as background.
\end{itemize}

Based on these, we compute the following metrics:
\begin{itemize}
    \item \textbf{Recall}: Measures the proportion of actual smoke pixels correctly
        identified. It is crucial for minimizing missed detections.
        \begin{equation}
            \text{Recall}= \frac{\text{TP}}{\text{TP} + \text{FN}}= \frac{|P \cap
            G|}{|G|}
        \end{equation}

    \item \textbf{Precision}: Measures the proportion of predicted smoke pixels that
        are correct.
        \begin{equation}
            \text{Precision}= \frac{\text{TP}}{\text{TP} + \text{FP}}= \frac{|P \cap
            G|}{|P|}
        \end{equation}

    \item \textbf{F1-Score}: The harmonic mean of Precision and Recall,
        providing a balanced measure.
        \begin{equation}
            \text{F1-Score}= 2 \times \frac{\text{Precision} \times
            \text{Recall}}{\text{Precision} + \text{Recall}}
        \end{equation}

    \item \textbf{Intersection over Union (IoU)} for the smoke class is defined
        as:
        \begin{equation}
            \text{IoU}_{\text{smoke}}= \frac{\text{TP}}{\text{TP} + \text{FP} +
            \text{FN}}= \frac{|P \cap G|}{|P \cup G|}
        \end{equation}
        The \textbf{Mean IoU (mIoU)} reported in our results is the average of
        $\text{IoU}_{\text{smoke}}$ and $\text{IoU}_{\text{background}}$, providing
        a standard measure of overall segmentation quality.
\end{itemize}

\subsection{Opacity-Specific Performance Metrics}
To assess performance on different smoke densities, we evaluate the binary smoke
prediction ($P$) against the distinct ground-truth opacity classes ($G_{high}$
and $G_{low}$).

\begin{itemize}
    \item \textbf{Opacity-level Recall ($\mathrm{Recall_{high}}, \mathrm{Recall_{low}}$)}:
        This measures the model's ability to detect smoke within a specific opacity
        class. For high-opacity smoke, it is defined as the fraction of ground-truth
        high-opacity pixels that were correctly predicted as smoke.
        \begin{equation}
            \mathrm{Recall_{high}}= \frac{|P \cap G_{high}|}{|G_{high}|}
        \end{equation}
        $\mathrm{Recall_{low}}$ is calculated analogously for $G_{low}$.

    \item \textbf{Opacity-level IoU ($\mathrm{IoU_{high}}, \mathrm{IoU_{low}}$)}:
        This metric is intended to measure the overlap between the predicted smoke
        pixels and the ground-truth pixels of a specific opacity. For high-opacity
        smoke, it is defined as:
        \begin{equation}
            \mathrm{IoU_{high}}= \frac{|P \cap G_{high}|}{|P \cup G_{high}|}
        \end{equation}
        $\mathrm{IoU_{low}}$ is calculated analogously. We acknowledge a potential
        fairness issue with this metric, as noted by reviewers. Since the model produces
        a single binary prediction $P$ for all smoke, the denominator
        $|P \cup G_{high}|$ includes all predicted smoke pixels, not just those intended
        to be high-opacity. This can artificially lower the score, especially if
        the model correctly identifies large areas of low-opacity smoke. However,
        this metric is retained for consistency and direct comparability with prior
        works that use the same formulation. It should be interpreted alongside
        Recall and F1-Score for a complete performance assessment.

    \item \textbf{Opacity-level F1-Score ($\mathrm{F1_{high}}, \mathrm{F1_{low}}$)}:
        To compute a balanced score for each opacity level, we first define an
        opacity-level precision. For high-opacity smoke, this is:
        \begin{equation}
            \mathrm{Precision_{high}}= \frac{|P \cap G_{high}|}{|P|}
        \end{equation}
        This measures what fraction of the total smoke prediction corresponds to
        the high-opacity class. The F1-score is then the harmonic mean of this
        precision and the corresponding recall.

    \item \textbf{Foreground mIoU ($\mathrm{mIoU_{smoke}}$)}: The average of $\mathrm{IoU_{high}}$
        and $\mathrm{IoU_{low}}$, providing a single, focused score for segmentation
        performance across the entire smoke opacity spectrum.
\end{itemize}




  \bibliographystyle{Latex/Classes/PhDbiblio-url2} 
  \renewcommand{\bibname}{References} 

  \bibliography{7_references/references} 








\end{document}